\documentclass[twoside,11pt]{article}
\usepackage{graphicx}
\usepackage{BOONDOX-cal} 
\usepackage{float} 
\usepackage{subcaption}
\usepackage{caption} 
\usepackage[ruled]{algorithm2e} 
\usepackage{xr} 
\usepackage{booktabs} 
\usepackage{multirow} 
\usepackage{color} 
\usepackage{setspace} 
\usepackage{amsfonts} 
\usepackage{amsmath} 
\usepackage{bm} 
\usepackage{longtable} 
\usepackage{enumitem} 
\usepackage{acronym}
\usepackage[a4paper,left=31.7mm,right=31.7mm,top=25.4mm,bottom=25.4mm]{geometry}

\SetKwComment{Comment}{(}{)}

\newcounter{mychapter}

\newtheorem{Def}{Definition}[section]

\title{\textbf{Outlyingness Scores with Cluster Catch Digraphs}}
\author{Rui Shi, Elvan Ceyhan, and Nedret Billor \\ Department of Mathematics and Statistics\\ Auburn University}
\date{}

\begin{document}
\maketitle

\begin{abstract}
\noindent
This paper introduces two novel, outlyingness scores (OSs) based on Cluster Catch Digraphs (CCDs):
\emph{Outbound Outlyingness Score} (OOS) and \emph{Inbound Outlyingness Score} (IOS).
These scores enhance the interpretability of outlier detection results. 
Both OSs employ graph-, density-, and distribution-based techniques, 
tailored to high-dimensional data with varying cluster shapes and intensities.
OOS evaluates the outlyingness of a point relative to its nearest neighbors, 
while IOS assesses the total ``influence" a point receives from others within its cluster.
Both OSs effectively identify global and local outliers, 
invariant to data collinearity.
Moreover, IOS is robust to the masking problems.
With extensive Monte Carlo simulations,
we compare the performance of both OSs with CCD-based, traditional, 
and state-of-the-art outlier detection methods.
Both OSs exhibit substantial overall improvements over the CCD-based methods in both artificial and real-world datasets,
particularly with IOS, 
which delivers the best overall performance among all the methods,
especially in high-dimensional settings.
\end{abstract}



\section{The Outlyingness Scores Based on Cluster Catch Digraphs}
\label{sec:CCD_Scores}
 
\subsection{Introduction and Motivation}
We have introduced several outlier detection methods based on RK-CCDs or UN-CCDs in the previous part, 
which share a common theme: 
(1) \textbf{Clustering Formation}: Identify potential clusters and construct a single or multiple covering balls around the center, covering majority points.
(2) \textbf{Outlier Identification:} Apply the \textbf{Density-based Mutual Catch Graph} (D-MCG) method within each cluster to isolate points distant from the center and label them as outliers. 
Evaluation results demonstrated the effectiveness of these methods, with the SUN-MCCD method exhibiting superior overall performance. 
However, the outlyingness is not a binary property \cite{breunig2000lof}.
Moreover, the CCD-based methods do not provide a measure of outlyingness. 
An \textbf{Outlyingness Score} (OS) quantifies the degree of outlyingness for each observation. Incorporating OSs offers several advantages \cite{kriegel2011interpreting}: 
(a) OSs enhance the interpretability of outlier detection results. 
(b) Ranking observations by their OSs enables user-defined thresholds for outlier classification. 
(c) OSs contribute to a richer understanding of the data distribution and point patterns.
Some established, well-known OS-based outlier detection methods include the following:
(i) \textbf{Isolation Forest} employs multiple decision trees to compute anomaly scores \cite{liu2008isolation}; 
(ii) \textbf{Local Information Graph-based Random Walk model} (LIGRW) identifies outliers by analyzing unusual patterns in random walks \cite{wang2018new};
(iii) \textbf{Local Outlier Factor} (LOF) is a prototype that utilizes \textbf{local reachability density} to compute outlyingness \cite{breunig2000lof};
(iv) \textbf{Connectivity-based Outlier Factor} (COF) enhances LOF by addressing limitations in detecting outliers with similar densities but differing neighbor patterns \cite{tang2002enhancing};
(v) \textbf{LOcal Correlation Integral} (LOCI) introduces a threshold-based score measuring the intensity of points within a given radius \cite{papadimitriou2003loci};
(vi) \textbf{Local Outlier Probabilities} (LoOP) assigns an outlier probability based on the deviates of a point from its local context \cite{kriegel2009loop};
(vii) \textbf{Outlier Detection using In-degree Number} (ODIN) identifies outliers as points with low in-degree numbers in a $k$NN graph \cite{hautamaki2004outlier}.
Other OS-based methods include \textbf{Angle-Based Outlier Detection} (ABOD) \cite{kriegel2008angle}, \textbf{Histogram-based Outlier Score} (HBOS) \cite{goldstein2012histogram}, and \textbf{Feature Bagging} \cite{lazarevic2005feature}.

We offer two novel, parameter-free methods to calculate OSs based on CCDs. 
These methods provide a quantitative measure of the degree to which an observation deviates from its neighbors.
The first approach, called \textbf{Outbound Outlyingness Score} (OOS), 
assesses the outlyingness of a point with its nearest neighbors; 
the second approach is \textbf{Inbound Outlyingness Score} (IOS), 
quantifies outlyingness as the inverse of the ``cumulative influence" a point receives from other members of its cluster. 
We will refer to them as \textbf{score-based (outlier detection) methods}.
Notably, both approaches offer a localized measure of outlyingness,
but they are also effective on global outliers.
Moreover, we show that IOS is robust to collective outliers exhibiting a masking effect.

\subsection{Outbound Outlyingness Score}

We first define \textbf{outbound neighbors} and the \textbf {vicinity density} as follows.

\begin{Def}[Outbound Neighbors]\label{def:ON}
Given a dataset $\mathrm{X} = \{x_1,x_2,...,x_{n}\}$ and a CCD constructed on it, 
the \textbf{outbound neighbors} of a point $x_i \in \mathrm{X}$, denoted as $N_O(x_i)$, are those points that are covered by the covering ball centered at $x_i$, i.e.,
\begin{equation}
  N_O(x_i):= \{x_j|x_j \in B(x_i,r_{x_i}), i \neq j\}.
\end{equation}
\end{Def}

It is worth noting $x_j \in N_O(X_i)$ does not necessarily imply $x_i \in N_O(X_j)$, 
which means this neighbor relationship is asymmetric. 
This is also why we call it ``outbound".

Introducing \textbf{vicinity density} around a point is one of the common ways to define an OS. 
Therefore, we define the vicinity density and an OS in the CCD context as follows.

\begin{Def}[Vicinity Density]
Given a dataset $\mathrm{X}$ and a point $x_i \in \mathrm{X}$, 
\textbf{the vicinity density} of $x_i$, 
denoted by $\rho_{x_i}$, is defined as follows,
\begin{equation}
  \rho_{x_i}:=\frac{|\{x_j|x_j \in B(x_i,r_{x_i})\}|}{r_{x_i}}^{\frac{1}{d}},
\end{equation}
where $|S|$ represent the cardinality of a set $S$, 
and $d$ is the dimensionality. 
\end{Def}

The quantity $\rho_{x_i}$ measures the point intensity of $B(x_i,r_{x_i})$.
If $x_i$ deviates from other points, 
$\rho_{x_i}$ should be small and much lower than those of $x_i$'s outbound neighbors (i.e., points in $N_O(x_i)$). 
Recall that a practical OS measures how different an observation is compared to other ``regular" points. 
With this notion, 
a point with a much lower vicinity density than its neighbors is more likely to be an outlier and should have a high OS. 
Therefore, we propose the following CCD-based OS.

\begin{Def}[Outbound Outlyingness Score (OOS)]
Given a dataset $\mathrm{X}$ and a point $x_i \in \mathrm{X}$, an \textbf{Outbound Outlyingness Score} of $x_i$, denoted as $OOS(x_i)$, is given as follows,
\begin{equation}
  OOS(x_i):=\frac{\sum_{x_j \in N_O(x_i)} \rho_{x_j}/|N_O(x_i)|}{\rho_{x_i}}.
\end{equation}
\end{Def}

The OOS of $x_i$ is the ratio between the vicinity densities of $x_i$ and its outbound neighbors.
Intuitively, for fixed $\rho_{x_i}$, 
the higher the average vicinity densities of $x_i$'s outbound neighbors, 
the higher $x_i$'s OOS, 
indicating a higher degree of outlyingness for $x_i$.

Initially, we considered defining the OS of a point $x_i$ as the reciprocal of the mean vicinity density of $N_O(x_i)$ (i.e., $\frac{1}{\sum_{x_j \in N_O(x_i)\rho_{x_j}/|N_O(x_i)|}}$), 
with the notion that if $x_i$'s outbound neighbors have high vicinity density, 
then the outlyingness degree of $x_i$ should be low. 
However, empirical evidence (not presented here) showed this approach is problematic because the outbound neighbors of $x_i$ may only consist of regular observations with high vicinity density, 
even if $x_i$ is an outlier. 
Therefore, we modified and improved the initial approach and proposed the OOS instead.
We demonstrate the mechanism of OOS method with an artificial dataset in the next section.

\subsubsection{An illustration of OOS with UN-CCDs}
\label{sec:OOS_Ex}

\begin{figure}[htb]
    \centering
    
    \begin{subfigure}[t]{0.48\textwidth}
        \includegraphics[width=\textwidth]{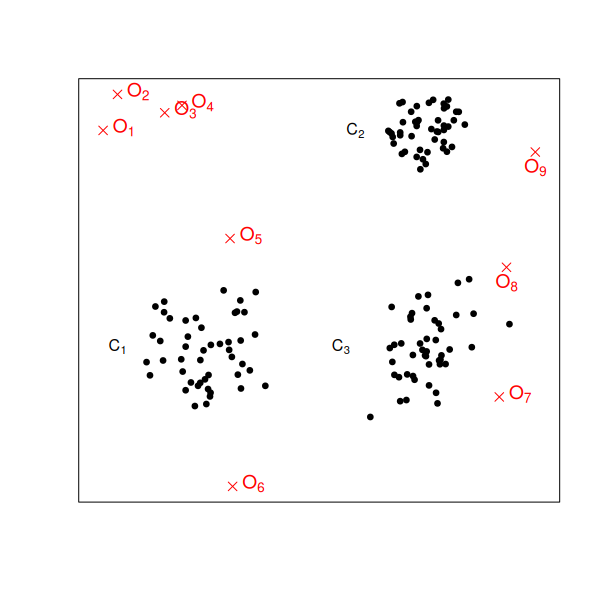}
        \caption{\large An artificial dataset}
        \label{fig:OOS1.1}
    \end{subfigure}
    \hfill 
    \begin{subfigure}[t]{0.48\textwidth}
        \includegraphics[width=\textwidth]{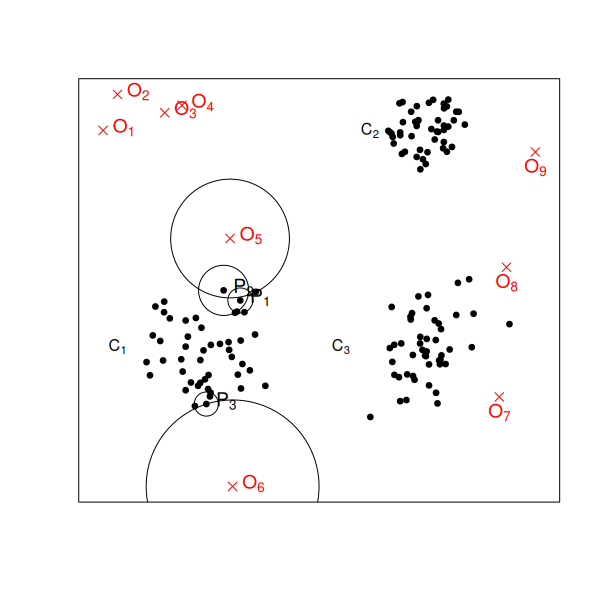}
        \caption{\large An illustration of OOS}
        \label{fig:OOS1.2}
    \end{subfigure}
    
    \vspace{1em} 
    
    \begin{subfigure}[t]{0.48\textwidth}
        \includegraphics[width=\textwidth]{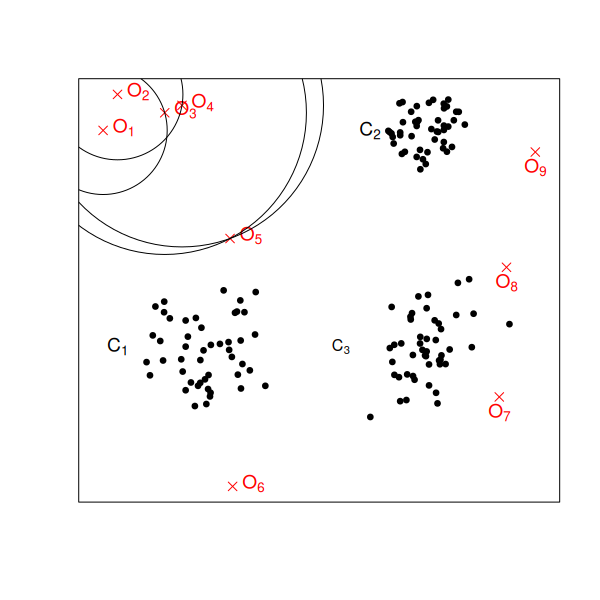}
        \caption{\large The masking problem}
        \label{fig:OOS1.3}
    \end{subfigure}
    \hfill 
    \begin{subfigure}[t]{0.48\textwidth}
        \includegraphics[width=\textwidth]{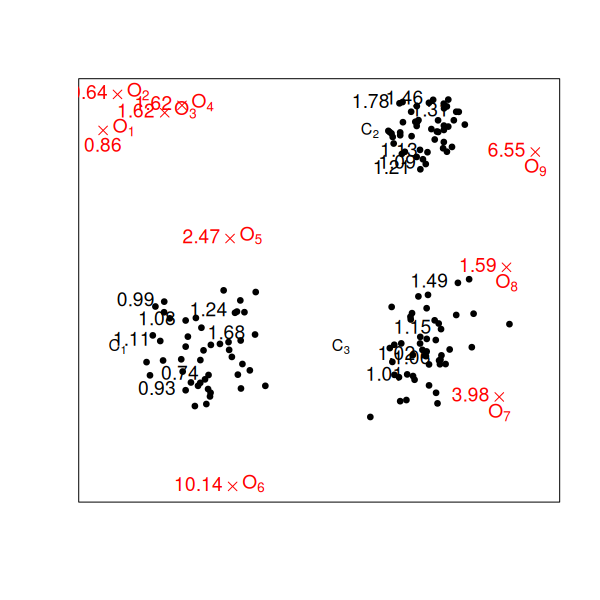}
        \caption{\large OOS values for the outliers}
        \label{fig:OOS1.4}
    \end{subfigure}
    
    \caption{An example of OOS on an artificial dataset with UN-CCDs. Black points are regular points and red crosses are outliers.}
    \label{fig:OOS1}
\end{figure}

To illustrate the efficacy of the OOS method in identifying outliers,
we use an artificial dataset (Figure \ref{fig:OOS1.1}) comprising three clusters, 
$C_1$, $C_2$, and $C_3$,
and nine outliers, $O_1$ to $O_9$ (highlighted in red).
Notably, $C_1$ and $C_2$ are drawn from uniform distributions on a disk,
with different densities,
and $C_3$ is generated from a Gaussian distribution with collinearity,
and the correlation is set to 0.5;
$O_1$ to $O_4$ are collective outliers, forming an outlier cluster,
while $O_5$ to $O_9$ are global (single) outliers far from regular points.

%

In Figure \ref{fig:OOS1.2}, 
we construct UN-CCD to illustrate the computation of OOS.
While OOS is independent of the clustering result, 
it is worth noting that the UN-CCD method assigns outliers $O_1, O_2,..., O_6$ to $C_1$, 
$O_7$ and $O_8$ to $C_3$,
and $O_9$ to $C_2$.
We first focus on the cluster $C_1$.
Figure \ref{fig:OOS1.2} presents the covering balls of $O_5$ and $O_6$, 
along with their outbound neighbors $P_i \ (i=1,2,3)$. 
$P_i's$ vicinity densities are substantially higher than those of $O_5$ and $O_6$, 
resulting in OOSs values of 2.47 and 10.14 for $O_5$ and $O_6$, respectively. 
In contrast, most of the remaining points within $C_1$ exhibit OOS values below 2 (Figure \ref{fig:OOS1.4}). 
Consequently, $O_5$ and $O_6$ are clearly distinguished by their markedly higher OOS values within $C_1$.
Similarly, $O_9$ receives a much higher OOS than the other points in $C_2$.

Consider $C_3$, although both $O_7$ and $O_8$ have similar distances to the cluster center,
$O_7$ should be more outlying due to the collinearity of cluster $C_3$.
The OOS method successfully captures this pattern, 
assigning $O_7$ a higher score.
Moreover, further empirical analysis (not presented here) shows that
OOS's performance remains stable with changing levels of collinearity.

Although OOS effectively captures global outliers, 
it is not robust to the masking problem, 
which is a common phenomenon and challenge in outlier detection,
and it occurs when a group of close outliers distorts local outlyingness calculations,
making it difficult to identify individual outliers within the group accurately.
Such a group of outliers is also referred to as \textbf{global or single outliers}.
 
Figure \ref{fig:OOS1.3} illustrates this problem, 
where outliers $O_1$, $O_2$, $O_3$, and $O_4$ are close, 
forming a small group on the top left, 
isolated from any other points.
Furthermore, they are outbound neighbors to each other,
excluding any regular points. 
This pattern leads to $O_1$ and $O_2$ having OOSs of 0.86 and 0.64, respectively, 
which are close to those of regular points.
Therefore, according to OOSs, 
we cannot distinguish $O_1$ and $O_2$ from other regular points.

In summary, OOS has the following advantages,
\begin{itemize}
  \item Effective on global and local (contextual) outliers.
  \item Robust to different levels of collinearity.
  \item OOSs are comparable globally since they do not depend on clustering.
\end{itemize}
However, as one of the limitations, OOS is not effective on collective outliers, showing little robustness to the masking problem.

To address this problem, 
we propose another OS, call \textbf{Inbound Outlyingness Score} (IOS).

\subsection{Inbound Outlyingness Score}

Unlike OOS, IOS evaluates the outlyingness by measuring the \textbf{cumulative influence} on a point, 
imposed by its \textbf{inbound neighbors},
which are defined as follows.

\begin{Def}[Inbound Neighbors] \label{def:Inbound_Neighbors}
Given a cluster $C=\{x_1,x_2,...,x_{n_c}\}$, 
where $n_c$ is the size of the cluster $C$, and a point $x_i \in C$, 
the \textbf{inbound neighbor} set of $x_i$, 
denoted as $N_I(x_i)$, are those points in $C$ whose covering balls covers $x_i$, i.e.,
\begin{equation}
  N_I(x_i):= \{x_j|x_i \in B(x_j,r_{x_j}), i \neq j\}.
\end{equation}
\end{Def}

It is worth noting that a point and its inbound neighbors are in the same cluster. 
Thus, inbound neighbors also depend on the clustering result of CCDs-based methods.
 
In the context of the RK-CCD and UN-CCD clustering methods,
suppose $x_i \in C$ is a point of interest.
The ball $B(x_j,r_{x_j})$ covering $x_i$ implies that
$x_i$ and $x_j$ belong to the same cluster. 
Therefore, the number of $x_i$'s inbound neighbors (i.e., the size of $N_I(x_i)$) can be interpreted as a measure of support for the inclusion of $x_i$ within a cluster.
Each inbound neighbor effectively acts as a ``vote", supporting $x_i$ being an inlier.
However, ``votes" are of different importance,
and we prioritize the ``votes" of points located at the denser regions of a cluster. 
Therefore, we set the vicinity density as the weight of each ``vote" and define the \textbf{cumulative influence} on $x_i$ as the sum of these weighted ``votes" from $N_I(x_i)$,
providing a refined metric for outlier detection.

\begin{Def}[Cumulative Influence] \label{def:CumInf}
Given a cluster $C$ (as in Definition \ref{def:Inbound_Neighbors}) and a point $x_i \in C$, 
the \textbf{cumulative influence} on $x_i$, denoted as $CI(x_i)$, is computed as follows,
\begin{equation}
  CI(x_i):= \sum_{x_j \in N_I(x_i)}\rho_{x_j}.
\end{equation}
\end{Def}

Note that the larger $CI(x_i)$, 
the less $x_i$ deviates from other points 
(i.e., the more likely $x_i$ belongs to the cluster). 
So, we may define the \textbf{Inbound Outlyingness Score} (IOS) of $x_i$ as the reciprocal of $CI(x_i)$. 
Furthermore, to avoid a zero denominator when $CI(x_i)$ is 0 (i.e., when a point does not have any inbound neighbors), 
we add $\rho(x_i)$ to $CI(x_i)$,
ranking the points without any inbound neighbors by the reciprocal of their vicinity densities.

\begin{Def}[Inbound Outlyingness Score (IOS)]
Given a cluster $C$ as before and a point $x_i \in C$, 
the Inbound Outlyingness Score (IOS) of $x_i$ is denoted as $IOS(x_i)$, is defined as:
\begin{equation}
  IOS(x_i):=\frac{1}{CI(x_i)+\rho(x_i)}.
\end{equation}
\end{Def}

We demonstrate the mechanism of IOS and its advantages over OOS with the same artificial dataset (Figure \ref{fig:OOS1}) in the next section.

\subsubsection{An illustration of IOS with UN-CCDs}
\label{sec:IOS_Ex}

\begin{figure}[htb]
    \centering
    
    \begin{subfigure}{0.48\textwidth}
        \includegraphics[width=\textwidth]{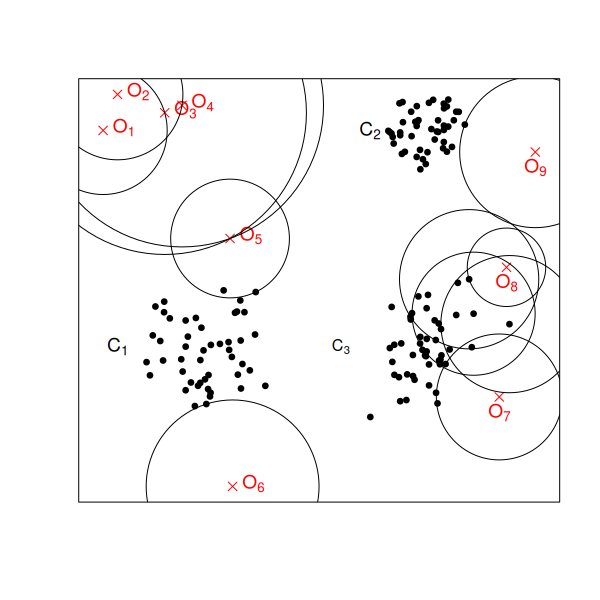}
        \caption{\large An illustration of IOS}
        \label{fig:IOS1.1}
    \end{subfigure}
    \hfill 
    \begin{subfigure}{0.48\textwidth}
        \includegraphics[width=\textwidth]{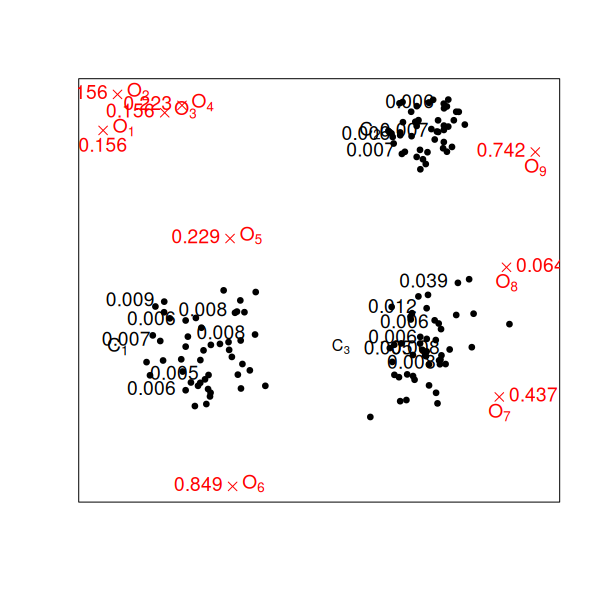}
        \caption{\large IOS values for the outliers}
        \label{fig:IOS1.2}
    \end{subfigure}
    
    \caption{An example of IOS with UN-CCDs on the same artificial dataset of Figure \ref{fig:OOS1.1}.}
    \label{fig:IOS1}
\end{figure}

Figure \ref{fig:IOS1.1} illustrates how IOS works on the same artificial dataset in Figure \ref{fig:OOS1}. 
We draw the covering balls of outliers and their inbound neighbors. 
The global outliers $O_6$, $O_7$, and $O_9$ do not have any inbound neighbors, 
and their IOSs are 0.849, 0.437, and 0.742, respectively, 
on the order of one hundred times larger than those of the regular points, 
which are typically around 0.007.
$O_5$'s IOS is 0.229, lower than those of $O_6$, $O_7$, 
and $O_9$ since $O_3$ and $O_4$ are its inbound neighbors.

Now consider the collective outliers $O_1$, $O_2$, $O_3$, and $O_4$,
which are close, 
forming a group of global outliers as shown in Figure \ref{fig:IOS1.1}. 
Moreover, they are inbound neighbors to each other. 
Nevertheless, they yield high IOSs because the cumulative influence from the other outliers is low. 
While OOS cannot distinguish $O_1$ and $O_2$ due to the masking problem, 
the IOSs of $O_1$ and $O_2$ are nearly 100 times than those of regular points, 
suggesting the robustness of IOS to the masking problems.

Consider the cluster $C_3$,
the IOS method can capture the collinearity pattern,
assigning $O_7$ a higher score than $O_8$ (0.437 vs 0.064).
similar to OOS, 
IOS is also invariant to different levels of collinearity, 
shown by further experimental analysis.


However, IOS is computed based on the inbound neighbors from the same cluster,
making it unsuitable for direct comparison between points in different clusters. 
This limitation arises because IOS is sensitive to cluster intensity variations, 
potentially providing misleading conclusions when ranking globally. 
Consider the outliers $O_6$ and $O_7$, belonging to $C_1$ and $C_2$ respectively, 
exhibiting similar distances to their respective clusters.
Therefore, $O_7$ should be more outlying than $O_6$ since $C_2$ has a significantly higher intensity than $C_1$.
However, the IOS of $O_6$ is higher, which is ``counter-intuitive".

To enable global IOS comparisons,
we standardize the IOS values. 
Moreover, standardization leads to a better interpretation of IOS. 
A common way of standardization is subtracting the sample mean ($\bar{X}_{IOS}$) and dividing by the sample standard deviation ($SD_{IOS}$):
\begin{equation}\label{equ:standarization}
  \frac{IOS(x_i)-\bar{X}_{IOS}}{SD_{IOS}}.
\end{equation}

By the ``three-sigma rule" \cite{maronna2019robust}, 
points with standardized scores larger than 3 can typically be deemed as outliers. 
However, this traditional measure is not robust to outliers, 
as the scores of outliers have a substantial and unbounded influence on the sample mean and the sample SD \cite{maronna2019robust}, 
which is particularly problematic in an outlier detection method.
Fortunately, there are robust alternatives to mean and SD, 
and we employ \textbf{median} (Med) and \textbf{the Normalized Median Absolute Deviation about the median} (MADN).
With this notion, a robust version of Equation \eqref{equ:standarization} can be defined as follows.

\begin{Def}
Given a cluster $C$ and a point $x_i \in C$, 
suppose the IOS values of points in $C$ are denoted as $IOS(C)$, 
then a robust standardization of $IOS(x_i)$ is given as follows, 
\begin{equation}\label{equ:Rstandarization}
  IOS^{std}(x_i) = \frac{IOS(x_i)-Med(IOS(C))}{MADN(IOS(C))},
\end{equation}
\end{Def}
where $MADN(IOS(C)) = Med\{IOS(C)-Med(IOS(C))\}/0.6745$. 
It is worth noting that the MADN of a random variable following $N(\mu, \delta)$ distribution is $\delta$ \cite{maronna2019robust}.
In the remaining sections, all IOS values will be standardized by default, i.e., $IOS(x_i)=IOS^{std}(x_i)$.

\begin{figure}[htb]
\centering
\includegraphics[width=0.6\textwidth]{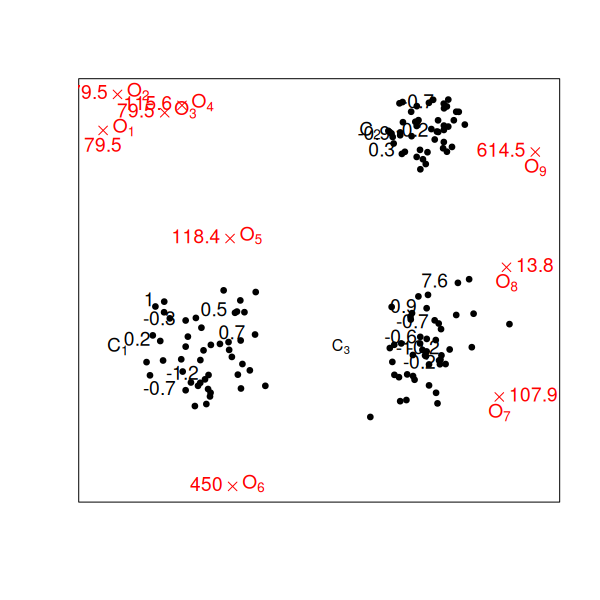}
\caption{Some standardized IOS values (rounded to one decimal point) with UN-CCDs on the same artificial dataset as Figure \ref{fig:OOS1.1}.}
\label{fig:IOS2}
\end{figure}

The standardized IOSs of some points are presented in Figure \ref{fig:IOS2}. 
The IOSs of most regular points are less than 1 (which could be negative),
much smaller than those of the outliers, which are at least 13.8. 
Moreover, $IOS(O_6)<IOS(O_7)$ after standardization, 
which aligns with people's intuition regarding their degree of outlyingness.

Observe that there are some ties between IOS values. 
For example, $IOS(O_1)=IOS(O_2)=IOS(O_3)=79.5$, 
because $O_1$, $O_2$, and $O_3$ share the same inbound neighbors. 
Generally, ties are not severe problems.
Nonetheless, we may need to break them in some instances. 
Therefore, we provide the following heuristic method to break ties according to vicinity density.

Suppose the IOSs of a cluster $C$ are ranked in ascending order as follows:
\begin{equation}\label{equ:IOS_ranking}
  IOS_{(1)} \leq ... \leq IOS_{(k-1)} \leq \underbrace{IOS_{(k)}=...=IOS_{(k)}}_{m} \leq IOS_{(k+m)} \leq ... \leq IOS_{(n_c)},
\end{equation}
where the IOS of, WLOG,  $x_1$, $x_2$, ..., $x_m$ equal to $IOS_{(k)}$ ($m\leq n_c$). 
To break ties, we assign new scores to them in a linearized fashion using $\rho_{x_1}$, $\rho_{x_2}$, ..., $\rho_{x_m}$. 
Without loss of generality, for any point $x_i$ ($i=1,...,m$), 
the standardized IOS of it, 
denoted as $\widetilde{IOS}(x_i)$, is defined as follows,
\begin{equation}\label{equ:B_ties}
  \widetilde{IOS}(x_i):=IOS_{(k+m)}-(IOS_{(k+m)}-IOS_{(k-1)})\frac{\rho_{x_i}}{\sum_{j}^{m}\rho_{x_j}}.
\end{equation}

After breaking the ties with Equation \eqref{equ:B_ties}, the new IOSs of $O_1$, $O_2$, and $O_3$ become 78.74, 73.89, and 94.07, 
the ordering of which is consistent with the ranking of their vicinity densities.

In summary, IOS has the following advantages,
\begin{itemize}
  \item Effective on both global and local (contextual) outliers.
  \item Effective on collective outliers and robust to the masking problem.
  \item Robust to the collinearity of data, and different levels of collinearity in the data.
  \item Although defined in a cluster-specific manner, IOSs are comparable globally after standardization.
\end{itemize}

We conclude this section by labeling these OSs. 
Both OOS and IOS depend on the digraph obtained from either RK-CCD or UN-CCD, 
resulting in four types of OOS, 
which are called \textbf{RKCCD-OOS}, \textbf{UNCCD-OOS}, \textbf{RKCCD-IOS}, and \textbf{UNCCD-IOS}, respectively. 
We assess the performance of these OS-based methods in the next section.

\section{Computational Complexity of OOS and IOS}
\label{sec:Complexity_OOS_IOS}

\paragraph{Time Complexity:}
After the initial CCDs have been constructed,
both the OOS and IOS take an additional $O(n^2)$ time at most,
dominated by the initial step of constructing the underlying CCDs.
We justify their additional time complexity as follows,
\begin{itemize}
  \item \textbf{Vicinity Density:} The first step is to calculate the vicinity density for each of the $n$ points. This requires iterating through the neighbor list of each point. This step takes at most $O(n^2)$ time.
  \item \textbf{OOS Calculation:} For each point, OOS finds its ``outbound neighbors" to compute the average vicinity density,
  and compares to the vicinity density of that point.
  which is at most $O(n^2)$ for the entire dataset.
  \item \textbf{IOS Calculation:} Calculating IOS requires summing the vicinity densities of all ``inbound neighbors" for each point. 
  This involves identifying which points' covering balls contain a given point. 
  This process, along with the subsequent standardization, also has a time complexity of at most $O(n^2)$.
\end{itemize}
Therefore, both OOS and IOS take an additional $O(n^2)$ time at most.

\paragraph{Space Complexity:} 
Similarly, the space complexity is determined by the underlying construction process of the CCD.
Both OOS and IOS require additional space for vicinity density and outlyingness scores, which take $O(n)$ space,
negligible in comparison to the $O(n^2)$ space taken by CCDs.
Therefore, the space complexity of OOS and IOS is $O(n^2)$.

\section{Monte Carlo Experiments with Synthetic Datasets}
\label{sec:Simulation_OS}

\subsection{General Settings}
We evaluate the performance of all the OSs by conducting Monte Carlo experiments with diverse simulation settings.
The simulation settings are elaborated in \cite{shi2024outlier}.
These settings involve different factors (e.g., dimensionality, dataset sizes, cluster volumes, etc.), 
which vary among datasets. 
Moreover, 
there are two major types of simulation settings, 
one with only uniform clusters and the other with only Gaussian clusters. 
The goal is to evaluate four newly proposed CCD-based OSs against four existing CCD-based methods \cite{shi2024outlier},
seeking evidence of performance improvements.

Preliminary simulations on artificial data identified ``elbow" cutoff points, serving as the thresholds for outlier detection. 
These thresholds, 
specific to each OS and dimensionality, 
are presented in Tables \ref{tab:Outlying_Score_NThreshold} and \ref{tab:Outlying_Score_GThreshold}. 
Notably, the datasets with Gaussian clusters require higher thresholds than those with uniform ones, 
because we want to differentiate outliers from regular data points around the tails of the Gaussian distributions.

\begin{table}[htb]
  \tiny
  \centering
  \footnotesize{\begin{tabular}{|c|c|c|c|c|c|c|c|}
  \hline
  & \multicolumn{7}{|c|}{Dimensionality $d$} \\ \cline{1-8}
  & $2$ & $3$ & $5$ & $10$ & $20$ & $50$ & $100$ \\ \cline{1-8}
  RKCCD-OOS & 6 & 6.5 & 5 & 4 & 4 & 14 & 13 \\ \cline{1-8}
  UNCCD-OOS & 4 & 4 & 4 & 3 & 3 & 5 & 13 \\ \cline{1-8}
  RKCCD-IOS & 4.5 & 4 & 4.5 & 5 & 4.5 & 6 & 7 \\ \cline{1-8}
  UNCCD-IOS & 6 & 4.5 & 4 & 3.5 & 4.5 & 3.5 & 6 \\ \cline{1-8}
  \end{tabular}}
  \caption{The thresholds for all the OSs when for datasets with only uniform clusters}\label{tab:Outlying_Score_NThreshold}
\end{table}

\begin{table}[htb]
  \tiny
  \centering
  \footnotesize{\begin{tabular}{|c|c|c|c|c|c|c|c|}
  \hline
  & \multicolumn{7}{|c|}{Dimensionality $d$} \\ \cline{1-8}
  & $2$ & $3$ & $5$ & $10$ & $20$ & $50$ & $100$ \\ \cline{1-8}
  RKCCD-OOS & 6 & 5.5 & 4.5 & 3.5 & 3.5 & 6.5 & 10 \\ \cline{1-8}
  UNCCD-OOS & 5.5 & 4.5 & 4 & 3.5 & 3 & 3 & 2.5 \\ \cline{1-8}
  RKCCD-IOS & 35 & 17 & 13 & 6.5 & 2.5 & 2.5 & 2.5 \\ \cline{1-8}
  UNCCD-IOS & 35 & 17 & 13 & 6.5 & 6 & 2.5 & 2.5 \\ \cline{1-8}
  \end{tabular}}
  \caption{The thresholds for all the OSs when for datasets with only Gaussian clusters}\label{tab:Outlying_Score_GThreshold}
\end{table}

\begin{figure}[htb]
    \centering
    
    \begin{subfigure}{1\textwidth}
        \includegraphics[width=\linewidth]{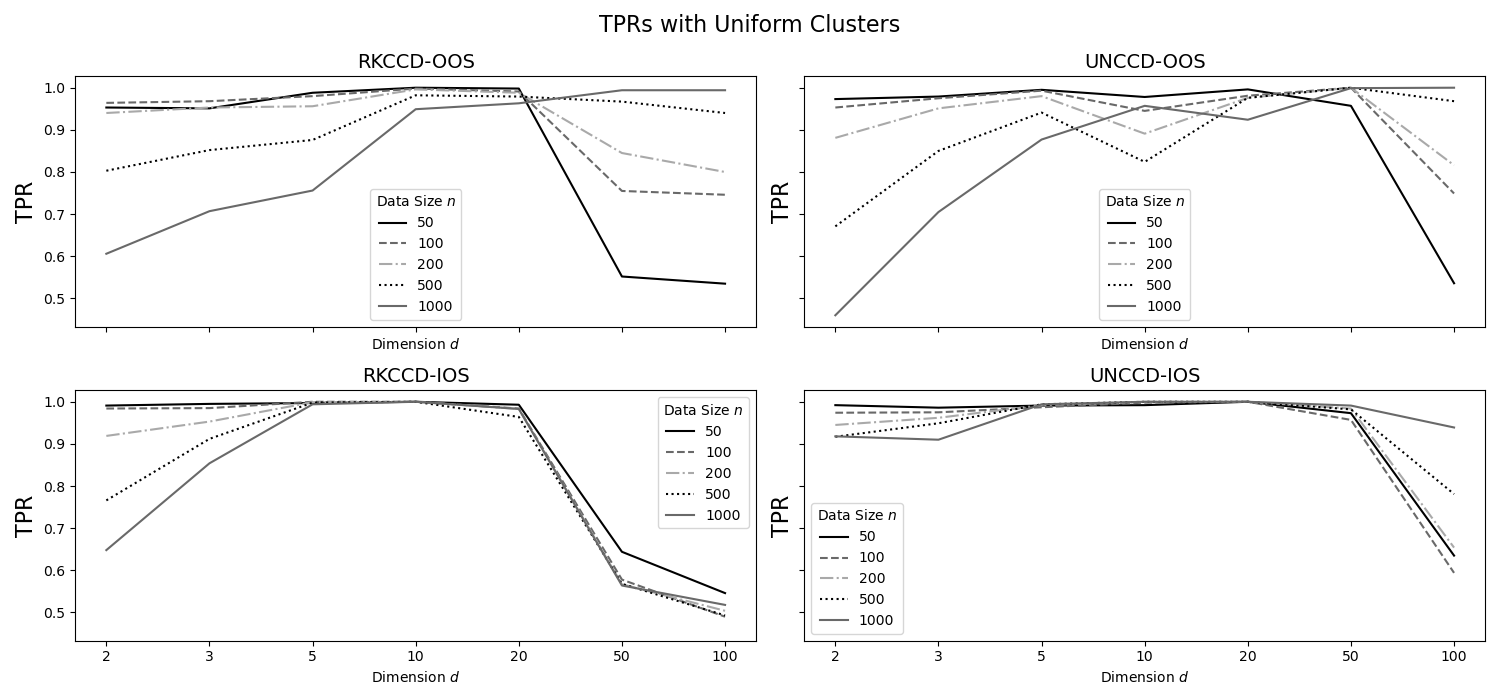}
    \end{subfigure}
    
    \begin{subfigure}{1\textwidth}
        \includegraphics[width=\linewidth]{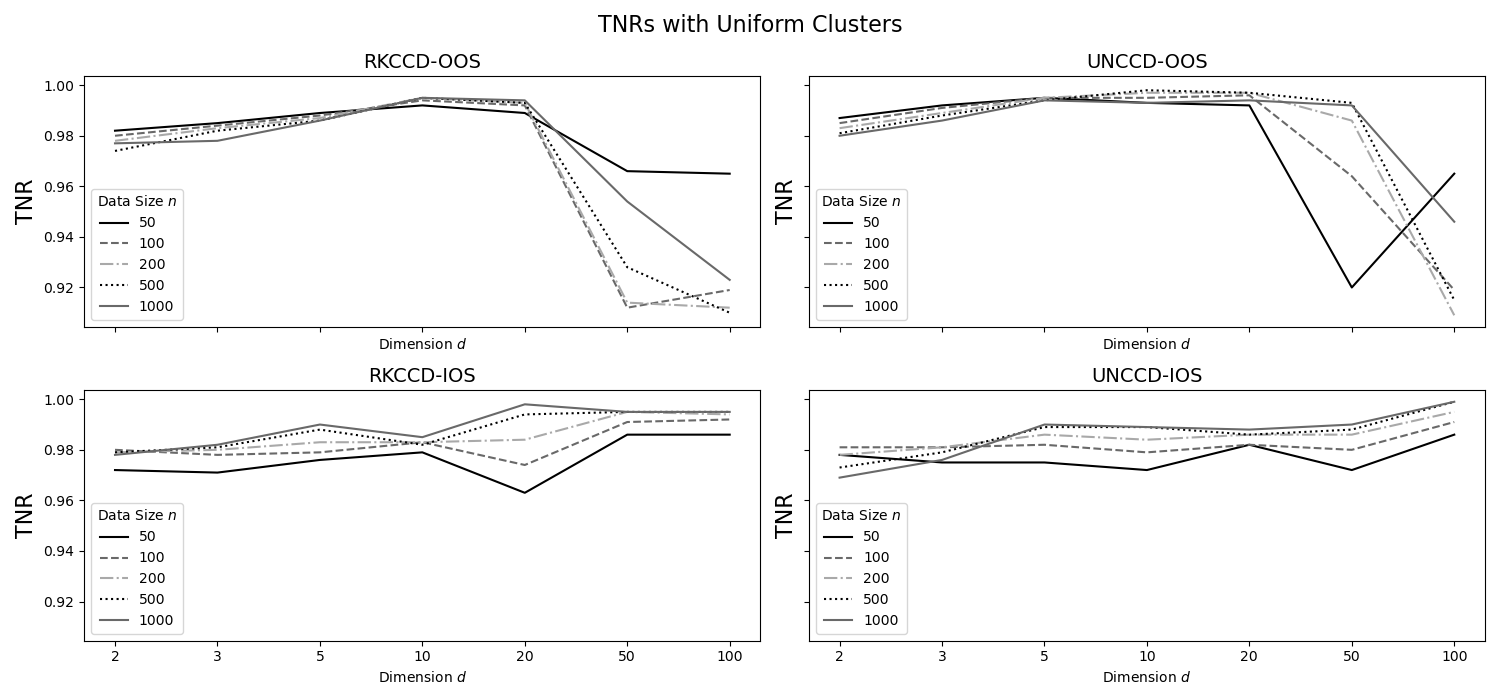}
    \end{subfigure}
    
    \caption{The line plots of the TPRs and TNRs of all CCD-based OSs, under the simulation settings (with uniform clusters) elaborated in \cite{shi2024outlier}.}
    \label{fig:Uniform_TPR_TNR_Lines_OS}
\end{figure}

\begin{figure}[htb]
    \centering
    
    \begin{subfigure}{1\textwidth}
        \includegraphics[width=\linewidth]{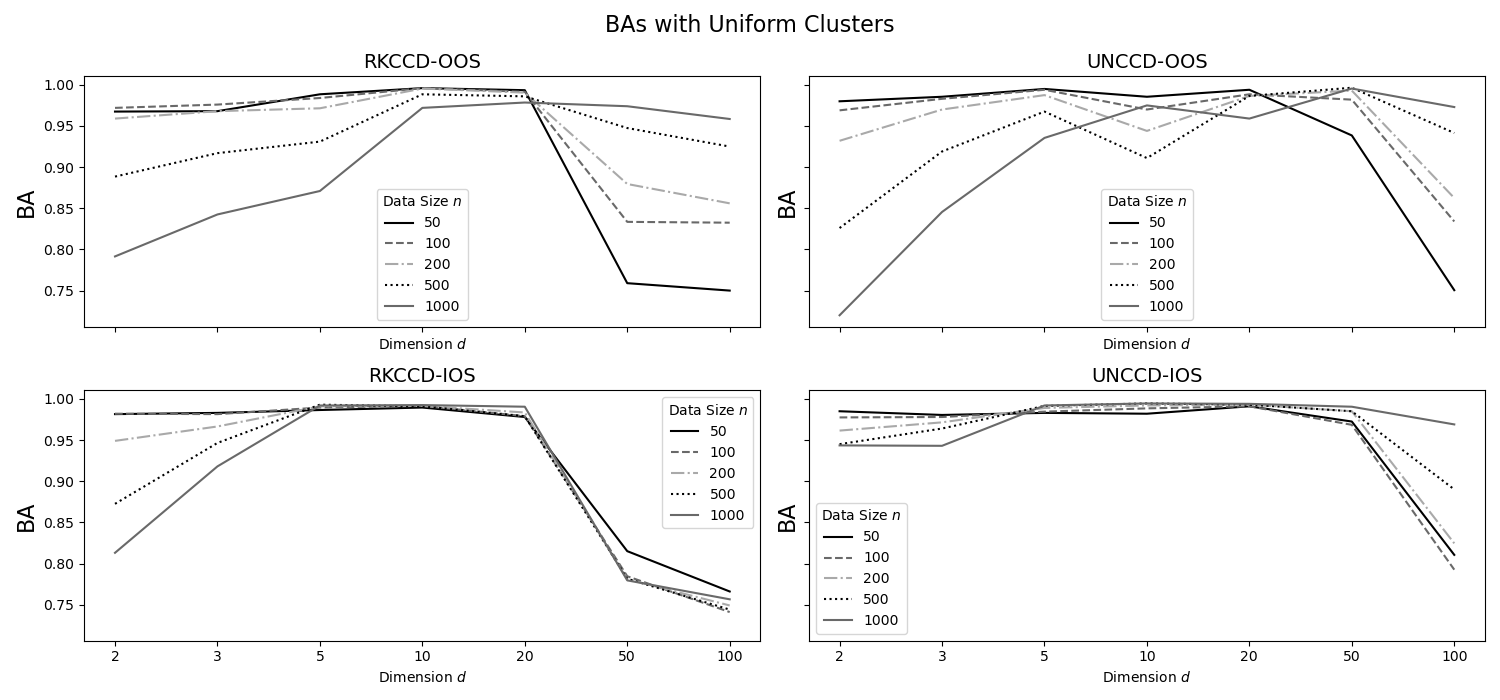}
    \end{subfigure}
    
    \begin{subfigure}{1\textwidth}
        \includegraphics[width=\linewidth]{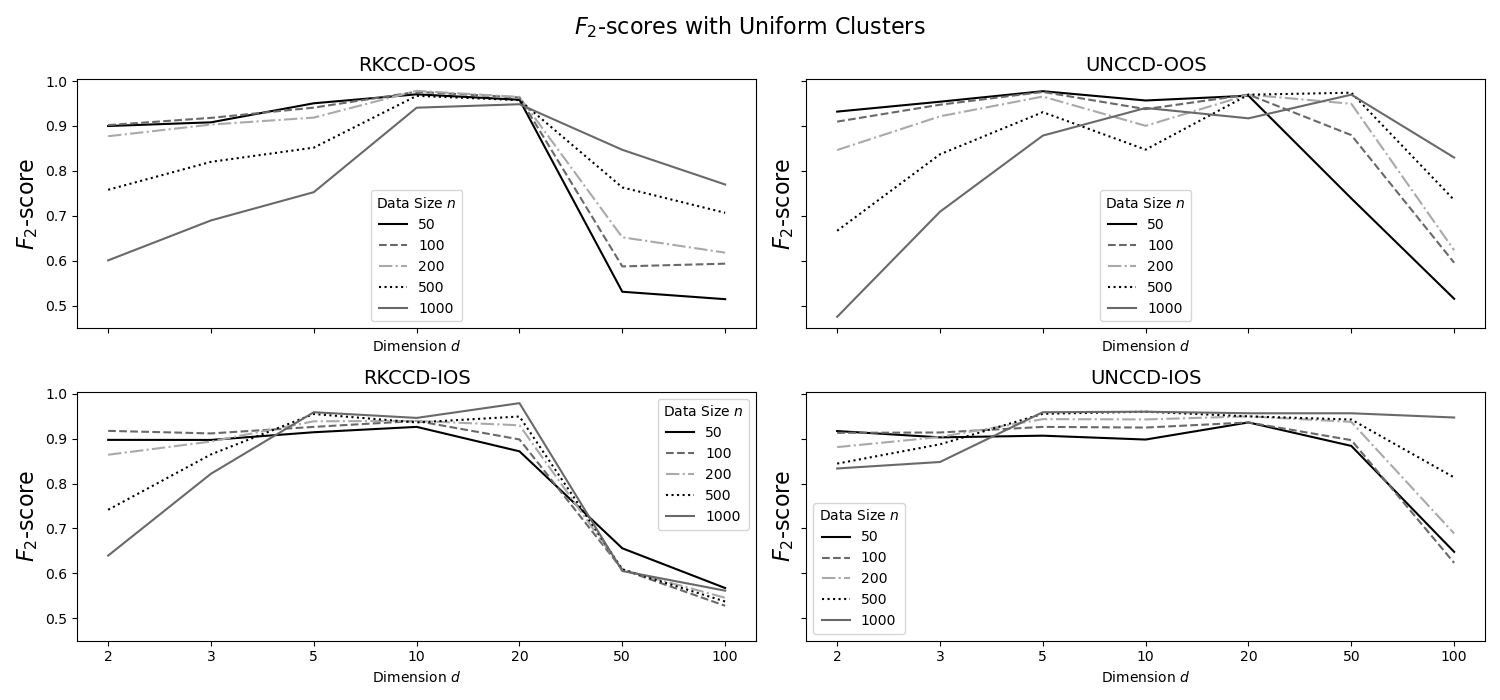}
    \end{subfigure}
    
    \caption{The line plots of the BAs and $F_2$-scores of all CCD-based OSs, under the simulation settings (with uniform clusters) elaborated in our previous work \cite{shi2024outlier}.}
    \label{fig:Uniform_BA_F_Lines_OS}
\end{figure}

\begin{figure}[htb]
    \centering
    
    \begin{subfigure}{1\textwidth}
        \includegraphics[width=\linewidth]{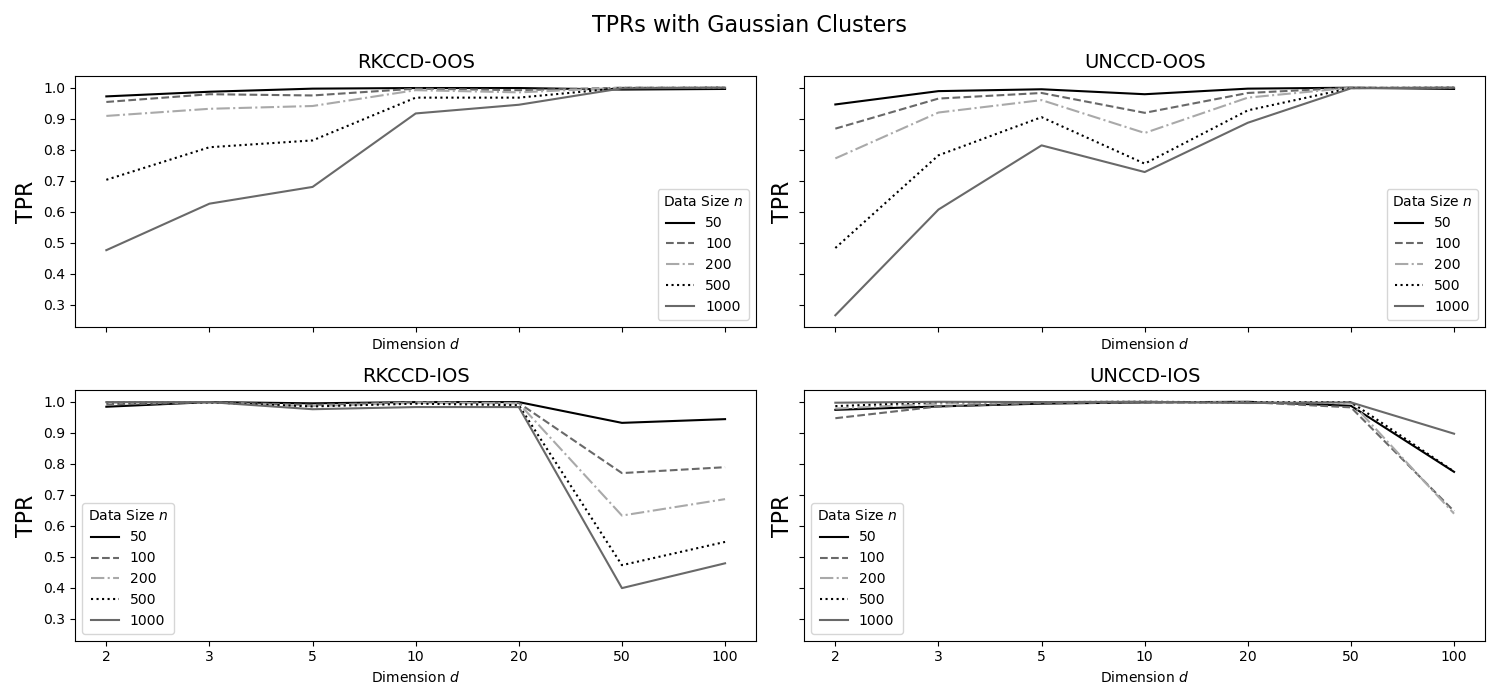}
    \end{subfigure}

    \begin{subfigure}{1\textwidth}
        \includegraphics[width=\linewidth]{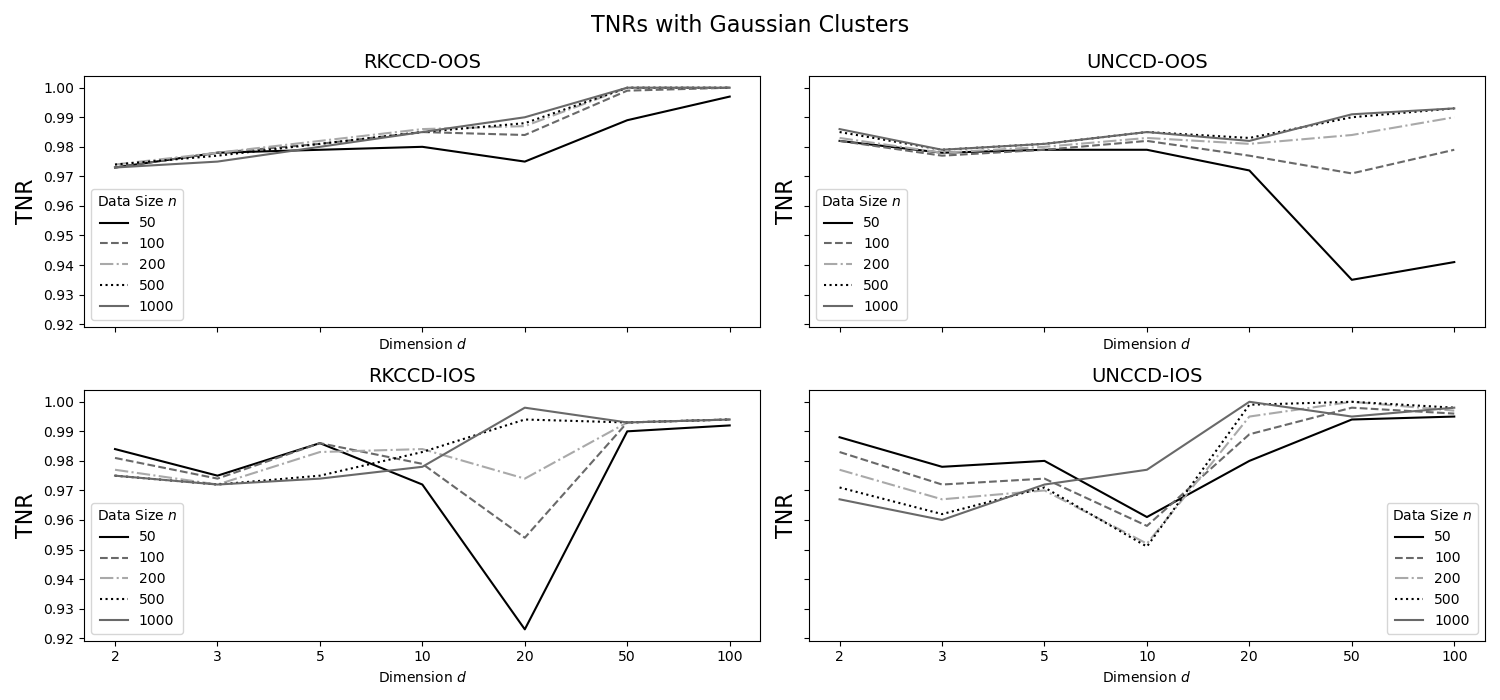}
    \end{subfigure}
    
    \caption{The line plots of the TPRs and TNRs of all CCD-based OSs, under the simulation settings (with Gaussian clusters) elaborated in our previous work \cite{shi2024outlier}.}
    \label{fig:Gaussian_TPR_TNR_Lines_OS}
\end{figure}

\begin{figure}[htb]
    \centering
    
    \begin{subfigure}{1\textwidth}
        \includegraphics[width=\linewidth]{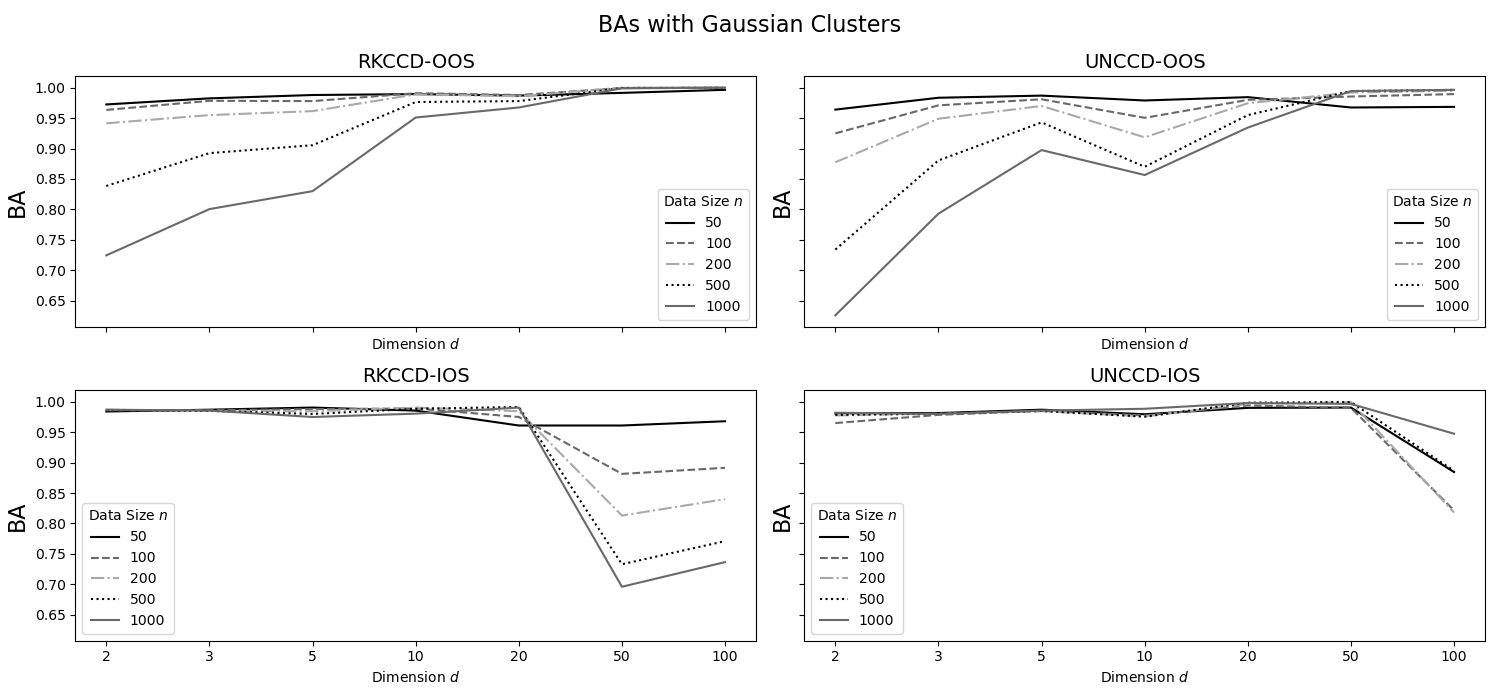}
    \end{subfigure}
    
    \begin{subfigure}{1\textwidth}
        \includegraphics[width=\linewidth]{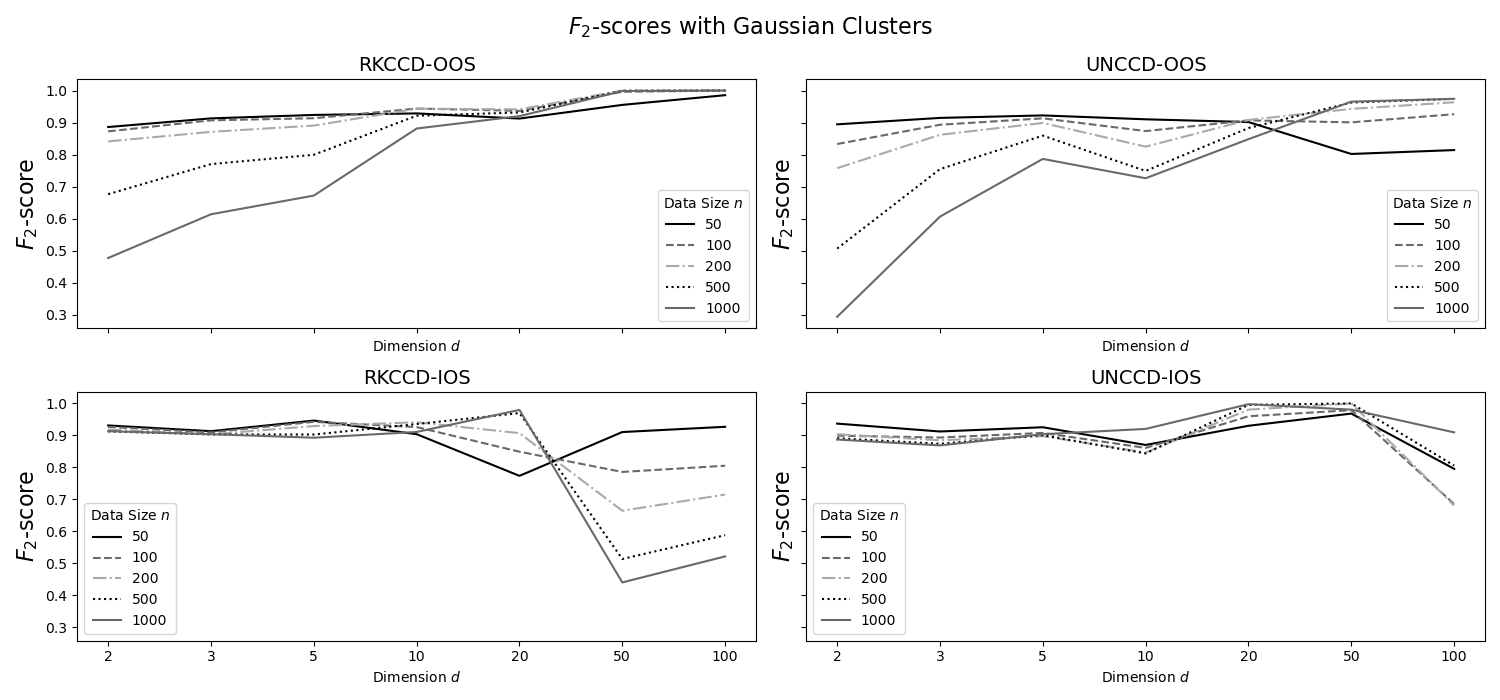}
    \end{subfigure}
    
    \caption{The line plots of the BAs and $F_2$-scores of all CCD-based OSs, under the simulation settings (with Gaussian clusters) elaborated in our previous work \cite{shi2024outlier}.}
    \label{fig:Gaussian_BA_F_Lines_OS}
\end{figure}

We select True Positive Rate (TPR), True Negative Rate (TNR), Balanced Accuracies (BA), and $F_2$-score as the performance metrics.
TPR is the proportion of outliers detected;
TNR is the proportion of regular points correctly labeled,
BA is the mean of TPR and TNR,
$F_{\beta}$-score is the weighted harmonic mean of recall (TPR) and precision, 
prioritizing positive observations (outliers) \cite{sasaki2007truth}. 
Since (in our opinion) recall is more important than precision in outlier detection, 
we set $\beta$ to 2, making recall twice as important.
We emphasize the latter two measures for enhanced accuracy since the size of outliers and regular points is highly imbalanced.
Comprehensive simulation results are provided in Tables \ref{tab:Uni_General_Results_OS1} to \ref{tab:Gau_General_Results_OS2}. 
For better visualization, we also provide the simulation results in four line plots (Figures \ref{fig:Uniform_TPR_TNR_Lines_OS} to \ref{fig:Gaussian_BA_F_Lines_OS}),
presenting the performance trend with varying dimensions and data sizes.

\begin{table}[ht]
  \centering
  \resizebox{\columnwidth}{!}{\begin{tabular}{|c|c|c|c|c|c|c|c|c|c|c|c|}
    \hline
    \multicolumn{2}{|c|}{} & \multicolumn{10}{|c|}{The Size of Datasets} \\ \cline{3-12}

    \multicolumn{2}{|c|}{} & \multicolumn{2}{|c|}{50} & \multicolumn{2}{|c|}{100} & \multicolumn{2}{|c|}{200} & \multicolumn{2}{|c|}{500} & \multicolumn{2}{|c|}{1000} \\ \cline{3-12}

    \multicolumn{2}{|c|}{} & TPR & TNR & TPR & TNR & TPR & TNR & TPR & TNR & TPR & TNR \\ \hline

    \multirow{4}*{$d=2$} & RKCCD-OOS & 0.953 & 0.982 & 0.964 & 0.980 & 0.940 & 0.978 & 0.796 & 0.977 & 0.606 & 0.977 \\ \cline{2-12}
    & UNCCD-OOS & 0.973 & 0.987 & 0.953 & 0.985 & 0.881 & 0.983 & 0.671 & 0.981 & 0.460 & 0.980 \\ \cline{2-12}
    & RKCCD-IOS & 0.991 & 0.972 & 0.984 & 0.980 & 0.919 & 0.979 & 0.779 & 0.979 & 0.648 & 0.978 \\ \cline{2-12}
    & UNCCD-IOS & 0.992 & 0.978 & 0.974 & 0.981 & 0.945 & 0.978 & 0.917 & 0.973 & 0.918 & 0.969 \\ \cline{1-12}

    \multirow{4}*{$d=3$} & RKCCD-OOS & 0.952 & 0.985 & 0.969 & 0.984 & 0.953 & 0.983 & 0.852 & 0.982 & 0.705 & 0.981 \\ \cline{2-12}
    & UNCCD-OOS & 0.979 & 0.992 & 0.975 & 0.991 & 0.951 & 0.989 & 0.850 & 0.988 & 0.705 & 0.986 \\ \cline{2-12}
    & RKCCD-IOS & 0.995 & 0.971 & 0.985 & 0.978 & 0.953 & 0.980 & 0.912 & 0.981 & 0.854 & 0.982 \\ \cline{2-12}
    & UNCCD-IOS & 0.986 & 0.975 & 0.975 & 0.981 & 0.962 & 0.981 & 0.949 & 0.979 & 0.910 & 0.976 \\ \cline{1-12}

    \multirow{4}*{$d=5$} & RKCCD-OOS & 0.988 & 0.989 & 0.980 & 0.988 & 0.956 & 0.987 & 0.876 & 0.986 & 0.756 & 0.986 \\ \cline{2-12}
    & UNCCD-OOS & 0.995 & 0.995 & 0.993 & 0.995 & 0.980 & 0.995 & 0.941 & 0.994 & 0.877 & 0.994 \\ \cline{2-12}
    & RKCCD-IOS & 0.997 & 0.976 & 1.000 & 0.979 & 0.999 & 0.983 & 0.998 & 0.988 & 0.994 & 0.990 \\ \cline{2-12}
    & UNCCD-IOS & 0.991 & 0.975 & 0.987 & 0.982 & 0.992 & 0.986 & 0.994 & 0.989 & 0.994 & 0.990 \\ \cline{1-12}

    \multirow{4}*{$d=10$} & RKCCD-OOS & 1.000 & 0.992 & 0.998 & 0.994 & 0.996 & 0.995 & 0.982 & 0.995 & 0.949 & 0.995 \\ \cline{2-12}
    & UNCCD-OOS & 0.978 & 0.993 & 0.945 & 0.995 & 0.891 & 0.997 & 0.824 & 0.998 & 0.957 & 0.993 \\ \cline{2-12}
    & RKCCD-IOS & 1.000 & 0.979 & 1.000 & 0.983 & 1.000 & 0.983 & 1.000 & 0.982 & 1.000 & 0.985 \\ \cline{2-12}
    & UNCCD-IOS & 0.992 & 0.972 & 0.998 & 0.979 & 1.000 & 0.984 & 1.000 & 0.989 & 1.000 & 0.989 \\ \cline{1-12}

    \multirow{4}*{$d=20$} & RKCCD-OOS & 0.998 & 0.989 & 0.991 & 0.992 & 0.988 & 0.993 & 0.979 & 0.993 & 0.963 & 0.994 \\ \cline{2-12}
    & UNCCD-OOS & 0.996 & 0.992 & 0.981 & 0.996 & 0.976 & 0.997 & 0.976 & 0.997 & 0.924 & 0.994 \\ \cline{2-12}
    & RKCCD-IOS & 0.993 & 0.963 & 0.984 & 0.974 & 0.983 & 0.984 & 0.964 & 0.994 & 0.983 & 0.998 \\ \cline{2-12}
    & UNCCD-IOS & 1.000 & 0.982 & 1.000 & 0.982 & 1.000 & 0.986 & 1.000 & 0.989 & 1.000 & 0.988 \\ \cline{1-12}

    \multirow{4}*{$d=50$} & RKCCD-OOS & 0.552 & 0.966 & 0.755 & 0.912 & 0.844 & 0.914 & 0.967 & 0.928 & 0.994 & 0.954 \\ \cline{2-12}
    & UNCCD-OOS & 0.957 & 0.920 & 1.000 & 0.964 & 1.000 & 0.986 & 1.000 & 0.993 & 0.999 & 0.992 \\ \cline{2-12}
    & RKCCD-IOS & 0.644 & 0.955 & 0.578 & 0.959 & 0.567 & 0.959 & 0.568 & 0.956 & 0.563 & 0.954 \\ \cline{2-12}
    & UNCCD-IOS & 0.973 & 0.972 & 0.957 & 0.980 & 0.984 & 0.986 & 0.982 & 0.988 & 0.991 & 0.990 \\ \cline{1-12}

    \multirow{4}*{$d=100$} & RKCCD-OOS & 0.535 & 0.965 & 0.746 & 0.919 & 0.800 & 0.912 & 0.940 & 0.910 & 0.994 & 0.923 \\ \cline{2-12}
    & UNCCD-OOS & 0.536 & 0.965 & 0.749 & 0.919 & 0.816 & 0.909 & 0.968 & 0.915 & 1.000 & 0.946 \\ \cline{2-12}
    & RKCCD-IOS & 0.546 & 0.986 & 0.490 & 0.992 & 0.504 & 0.994 & 0.493 & 0.995 & 0.518 & 0.995 \\ \cline{2-12}
    & UNCCD-IOS & 0.635 & 0.986 & 0.594 & 0.991 & 0.654 & 0.995 & 0.781 & 0.999 & 0.939 & 0.999 \\ \cline{1-12}
  \end{tabular}}
  \caption{Summary of the TPRs and TNRs of all the CCD-based OSs, with the simulation settings (with uniform clusters) elaborated in \cite{shi2024outlier}.}\label{tab:Uni_General_Results_OS1}
\end{table}


\begin{table}[ht]
  \centering
  \resizebox{\columnwidth}{!}{\begin{tabular}{|c|c|c|c|c|c|c|c|c|c|c|c|}
    \hline
    \multicolumn{2}{|c|}{} & \multicolumn{10}{|c|}{The Size of Datasets} \\ \cline{3-12}

    \multicolumn{2}{|c|}{} & \multicolumn{2}{|c|}{50} & \multicolumn{2}{|c|}{100} & \multicolumn{2}{|c|}{200} & \multicolumn{2}{|c|}{500} & \multicolumn{2}{|c|}{1000} \\ \cline{3-12}

    \multicolumn{2}{|c|}{} & BA & $F_2$-score & BA & $F_2$-score & BA & $F_2$-score & BA & $F_2$-score & BA & $F_2$-score \\ \hline

    \multirow{4}*{$d=2$} & RKCCD-OOS & 0.968 & 0.900 & 0.972 & 0.902 & 0.959 & 0.877 & 0.887 & 0.761 & 0.792 & 0.601 \\ \cline{2-12}
    & UNCCD-OOS & 0.980 & 0.932 & 0.969 & 0.910 & 0.932 & 0.846 & 0.826 & 0.667 & 0.720 & 0.475 \\ \cline{2-12}
    & RKCCD-IOS & 0.982 & 0.897 & 0.982 & 0.917 & 0.949 & 0.864 & 0.879 & 0.752 & 0.813 & 0.640 \\ \cline{2-12}
    & UNCCD-IOS & 0.985 & 0.917 & 0.978 & 0.913 & 0.962 & 0.881 & 0.945 & 0.844 & 0.944 & 0.833 \\ \cline{1-12}

    \multirow{4}*{$d=3$} & RKCCD-OOS & 0.968 & 0.908 & 0.976 & 0.918 & 0.968 & 0.903 & 0.917 & 0.820 & 0.843 & 0.696 \\ \cline{2-12}
    & UNCCD-OOS & 0.986 & 0.954 & 0.983 & 0.947 & 0.970 & 0.922 & 0.919 & 0.837 & 0.846 & 0.709 \\ \cline{2-12}
    & RKCCD-IOS & 0.983 & 0.897 & 0.982 & 0.912 & 0.967 & 0.893 & 0.947 & 0.865 & 0.918 & 0.822 \\ \cline{2-12}
    & UNCCD-IOS & 0.981 & 0.903 & 0.978 & 0.914 & 0.972 & 0.904 & 0.964 & 0.887 & 0.943 & 0.848 \\ \cline{1-12}

    \multirow{4}*{$d=5$} & RKCCD-OOS & 0.989 & 0.951 & 0.984 & 0.941 & 0.972 & 0.919 & 0.931 & 0.852 & 0.871 & 0.753 \\ \cline{2-12}
    & UNCCD-OOS & 0.995 & 0.977 & 0.994 & 0.976 & 0.988 & 0.966 & 0.968 & 0.931 & 0.936 & 0.879 \\ \cline{2-12}
    & RKCCD-IOS & 0.987 & 0.914 & 0.990 & 0.926 & 0.991 & 0.939 & 0.993 & 0.955 & 0.992 & 0.959 \\ \cline{2-12}
    & UNCCD-IOS & 0.983 & 0.907 & 0.985 & 0.926 & 0.989 & 0.943 & 0.992 & 0.955 & 0.992 & 0.959 \\ \cline{1-12}

    \multirow{4}*{$d=10$} & RKCCD-OOS & 0.996 & 0.970 & 0.996 & 0.976 & 0.996 & 0.978 & 0.989 & 0.967 & 0.972 & 0.941 \\ \cline{2-12}
    & UNCCD-OOS & 0.986 & 0.957 & 0.970 & 0.938 & 0.944 & 0.900 & 0.911 & 0.847 & 0.975 & 0.940 \\ \cline{2-12}
    & RKCCD-IOS & 0.990 & 0.926 & 0.992 & 0.939 & 0.992 & 0.939 & 0.991 & 0.936 & 0.993 & 0.946 \\ \cline{2-12}
    & UNCCD-IOS & 0.982 & 0.898 & 0.989 & 0.925 & 0.992 & 0.943 & 0.995 & 0.960 & 0.995 & 0.960 \\ \cline{1-12}

    \multirow{4}*{$d=20$} & RKCCD-OOS & 0.994 & 0.958 & 0.992 & 0.963 & 0.991 & 0.965 & 0.986 & 0.958 & 0.979 & 0.948 \\ \cline{2-12}
    & UNCCD-OOS & 0.994 & 0.967 & 0.989 & 0.970 & 0.987 & 0.970 & 0.987 & 0.970 & 0.959 & 0.917 \\ \cline{2-12}
    & RKCCD-IOS & 0.978 & 0.872 & 0.979 & 0.898 & 0.984 & 0.930 & 0.979 & 0.949 & 0.991 & 0.979 \\ \cline{2-12}
    & UNCCD-IOS & 0.991 & 0.936 & 0.991 & 0.936 & 0.993 & 0.946 & 0.993 & 0.949 & 0.994 & 0.956 \\ \cline{1-12}

    \multirow{4}*{$d=50$} & RKCCD-OOS & 0.759 & 0.531 & 0.834 & 0.587 & 0.880 & 0.652 & 0.948 & 0.763 & 0.974 & 0.847 \\ \cline{2-12}
    & UNCCD-OOS & 0.939 & 0.739 & 0.982 & 0.880 & 0.993 & 0.949 & 0.997 & 0.974 & 0.996 & 0.970 \\ \cline{2-12}
    & RKCCD-IOS & 0.815 & 0.656 & 0.785 & 0.609 & 0.781 & 0.608 & 0.782 & 0.609 & 0.780 & 0.605 \\ \cline{2-12}
    & UNCCD-IOS & 0.973 & 0.884 & 0.969 & 0.897 & 0.985 & 0.937 & 0.985 & 0.942 & 0.991 & 0.956 \\ \cline{1-12}

    \multirow{4}*{$d=100$} & RKCCD-OOS & 0.750 & 0.514 & 0.833 & 0.593 & 0.856 & 0.618 & 0.925 & 0.707 & 0.959 & 0.770 \\ \cline{2-12}
    & UNCCD-OOS & 0.751 & 0.515 & 0.834 & 0.596 & 0.863 & 0.623 & 0.942 & 0.735 & 0.973 & 0.830 \\ \cline{2-12}
    & RKCCD-IOS & 0.766 & 0.567 & 0.741 & 0.528 & 0.749 & 0.546 & 0.744 & 0.537 & 0.757 & 0.561 \\ \cline{2-12}
    & UNCCD-IOS & 0.811 & 0.648 & 0.793 & 0.623 & 0.825 & 0.689 & 0.890 & 0.814 & 0.969 & 0.947 \\ \cline{1-12}
  \end{tabular}}
  \caption{Summary of the BAs and $F_2$-scores of all the CCD-based OSs, with the simulation settings (with uniform clusters) elaborated in \cite{shi2024outlier}.}\label{tab:Uni_General_Results_OS2}
\end{table}


Firstly, we consider the simulation results with uniform clusters (Tables \ref{tab:Uni_General_Results_OS1} and \ref{tab:Uni_General_Results_OS2}). 
We will begin by exploring the behaviors of the four CCD-based OSs, 
and compare them with the previously proposed CCD-based outlier detection methods.

We first look at RKCCD-OOS and UNCCD-OOS. 
They exhibit similar performance across different dimensions.
When $d \leq 20$, both OOSs achieve high BAs and $F_2$-scores,
often exceeding 0.95 and 0.90, respectively, 
when the size of the dataset ($n$) is less than or equals to 200. 
However, as $n$ exceeds 500, 
the masking problem arises due to the increasing number of outliers. 
This issue is particularly prominent in lower dimensions where outlier intensities are high. 
For instance, with $d=2$ and $n=50$, UNCCD-OOS attains a BA of 0.980 and an $F_2$-scores of 0.932, 
but they drop to 0.720 and 0.475, respectively, as $n$ reaches 1000. 
Conversely, in high-dimensional settings ($d \geq 50$), 
both OOSs achieve higher TPRs with larger datasets. 
For instance, when $d=100$, the TPR of RKCCD-OOS progressively increases from 0.535 to 0.994 as $n$ increases. 
This trend can be attributed to the fact that in high-dimensional spaces with limited data, 
the vicinity densities of regular observations and outliers may not differ substantially. 
Recall that OOS measures the relative vicinity density of a point with its outbound neighbors, 
the reduced performance with smaller datasets is expected. 
Furthermore, when $d=50$ and $n=50$, the TPR of UNCCD-OOS is 0.957, substantially higher than that of RKCCD-OOS (0.535). 
This is because the covering balls of UN-CCDs are typically larger than those of RK-CCD, 
resulting in considerable differences between the vicinity densities between regular observations and outliers, 
which ultimately enhances the performance of OOS.

Second, we evaluate the performance of RKCCD-IOS and UNCCD-IOS. 
With $d \leq 20$, both IOSs demonstrate comparable performance to OOSs when the data size $n$ is small.
Moreover, an improvement is observed with larger $n$ values when compared to the two OOSs. 
For instance, with $d=3$ and $n\leq200$, UNCCD-IOS achieves $F_2$-scores comparable to those of UNCCD-OOS (0.903, 0.914, and 0.904 versus 0.954, 0.947, and 0.922). 
However, when $n\geq500$, UNCCD-IOS shows substantially higher $F_2$-scores than those of UNCCD-OOS (0.887, 0.848 versus 0.837, 0.709). 
This stems from the inherent advantage of IOS, 
which measures the cumulative influence a point receives from other points, 
and the cumulative influence of outliers is typically low compared to regular observations, 
making IOS more robust against the masking problem, 
particularly in simulation settings with many outliers. 
For example, with $d=2$ and $n=1000$, 
the $F_2$-scores of UNCCD-OOS and UNCCD-IOS are 0.475 and 0.833, respectively, showing a huge difference.
However, RKCCD-IOS does not perform well when $d\geq50$. 
For example, when $d=50$ and $n=1000$, the $F_2$-score of RKCCD-IOS is merely 0.605, 
much lower than that of RKCCD-OOS, which is 0.847. 
A reasonable explanation is the inherent challenges of high-dimensional clustering. 
As $d$ increases, 
the average distances between outliers and clusters grow, 
and the covering balls constructed by RK-CCDs are not large enough for high dimensions \cite{shi2024outlier}. 
Consequently, as the number of outliers increases, 
RK-CCDs may falsely identify a set of close outliers as a valid cluster since they are not robust to the masking problem. 
Furthermore, IOS identifies an outlier by within-cluster comparisons, which does not mitigate this problem.

Next, we compare the simulation results of the CCD-based OSs with previously proposed CCD-based outlier detection methods (which can be found in our previous work \cite{shi2024outlier}).

First, we consider the U-MCCD method, RKCCD-OOS, and RKCCD-IOS. 
When $d \leq 5$, the U-MCCD method and RKCCD-IOS exhibit comparable performance, 
and RKCCD-OOS performs worse when $n$ is large due to the masking problem. 
All three methods demonstrate high efficacy at $d=10$, 
achieving high $F_2$-scores.
When $d \geq 20$, the performance of the U-MCCD method drops substantially, 
while both RKCCD-OOS and RKCCD-IOS maintain better performance, achieving higher $F_2$-scores.

Then, we compare the UNCCD-OOS, UNCCD-IOS, and UN-MCCD methods. 
When $d \leq 20$, the UN-MCCD method performs comparably or better than the two OSs, 
especially when $d$ is small and $n$ is large. 
UNCCD-OOS consistently delivers the weakest performance because of the masking problem. 
However, a significant performance shift occurs when $d \geq 50$. 
The UN-MCCD method degrades substantially, 
with $F_2$-scores falling below 0.5 when $d=50$ and below 0.3 when $d=50$. 
In contrast, both OSs exhibit substantial improvement, 
especially UNCCD-IOS, 
whose $F_2$-scores are at least 0.85 when $d=50$.


Following the analysis of simulations with uniform clusters, 
we explore the results obtained from simulations with Gaussian clusters (see Tables \ref{tab:Gau_General_Results_OS1} and \ref{tab:Gau_General_Results_OS2}). 
Similarly, we begin by assessing the performance of the four OSs, 
and then compare them with the previously proposed CCD-based outlier detection methods (see our previous work \cite{shi2024outlier}).

\begin{table}[ht]
  \centering
  \resizebox{\columnwidth}{!}{\begin{tabular}{|c|c|c|c|c|c|c|c|c|c|c|c|}
    \hline
    \multicolumn{2}{|c|}{} & \multicolumn{10}{|c|}{The Size of Datasets} \\ \cline{3-12}

    \multicolumn{2}{|c|}{} & \multicolumn{2}{|c|}{50} & \multicolumn{2}{|c|}{100} & \multicolumn{2}{|c|}{200} & \multicolumn{2}{|c|}{500} & \multicolumn{2}{|c|}{1000} \\ \cline{3-12}

    \multicolumn{2}{|c|}{} & TPR & TNR & TPR & TNR & TPR & TNR & TPR & TNR & TPR & TNR \\ \hline

    \multirow{4}*{$d=2$} & RKCCD-OOS & 0.972 & 0.973 & 0.954 & 0.973 & 0.909 & 0.974 & 0.704 & 0.974 & 0.475 & 0.974 \\ \cline{2-12}
    & UNCCD-OOS & 0.946 & 0.982 & 0.868 & 0.982 & 0.772 & 0.983 & 0.483 & 0.985 & 0.266 & 0.986 \\ \cline{2-12}
    & RKCCD-IOS & 0.984 & 0.984 & 0.991 & 0.981 & 0.997 & 0.977 & 0.998 & 0.975 & 0.999 & 0.975 \\ \cline{2-12}
    & UNCCD-IOS & 0.974 & 0.988 & 0.947 & 0.983 & 0.978 & 0.977 & 0.986 & 0.971 & 0.997 & 0.967 \\ \cline{1-12}

    \multirow{4}*{$d=3$} & RKCCD-OOS & 0.987 & 0.978 & 0.979 & 0.978 & 0.932 & 0.978 & 0.808 & 0.977 & 0.626 & 0.975 \\ \cline{2-12}
    & UNCCD-OOS & 0.989 & 0.978 & 0.965 & 0.977 & 0.920 & 0.978 & 0.782 & 0.979 & 0.607 & 0.979 \\ \cline{2-12}
    & RKCCD-IOS & 0.999 & 0.975 & 0.998 & 0.974 & 0.998 & 0.972 & 0.998 & 0.972 & 0.999 & 0.972 \\ \cline{2-12}
    & UNCCD-IOS & 0.985 & 0.978 & 0.985 & 0.972 & 0.994 & 0.967 & 0.999 & 0.962 & 1.000 & 0.960 \\ \cline{1-12}

    \multirow{4}*{$d=5$} & RKCCD-OOS & 0.997 & 0.979 & 0.975 & 0.981 & 0.941 & 0.982 & 0.830 & 0.981 & 0.680 & 0.980 \\ \cline{2-12}
    & UNCCD-OOS & 0.995 & 0.979 & 0.983 & 0.979 & 0.960 & 0.980 & 0.905 & 0.981 & 0.814 & 0.981 \\ \cline{2-12}
    & RKCCD-IOS & 0.995 & 0.986 & 0.990 & 0.986 & 0.986 & 0.983 & 0.985 & 0.975 & 0.976 & 0.974 \\ \cline{2-12}
    & UNCCD-IOS & 0.994 & 0.980 & 0.996 & 0.974 & 0.999 & 0.970 & 0.999 & 0.971 & 0.999 & 0.972 \\ \cline{1-12}

    \multirow{4}*{$d=10$} & RKCCD-OOS & 0.999 & 0.980 & 0.997 & 0.985 & 0.992 & 0.986 & 0.968 & 0.985 & 0.917 & 0.985 \\ \cline{2-12}
    & UNCCD-OOS & 0.979 & 0.979 & 0.919 & 0.982 & 0.854 & 0.983 & 0.755 & 0.985 & 0.728 & 0.985 \\ \cline{2-12}
    & RKCCD-IOS & 0.999 & 0.972 & 0.999 & 0.979 & 0.997 & 0.984 & 0.994 & 0.983 & 0.983 & 0.978 \\ \cline{2-12}
    & UNCCD-IOS & 0.998 & 0.961 & 0.997 & 0.958 & 1.000 & 0.952 & 1.000 & 0.951 & 1.000 & 0.963 \\ \cline{1-12}

    \multirow{4}*{$d=20$} & RKCCD-OOS & 0.999 & 0.975 & 0.992 & 0.984 & 0.985 & 0.987 & 0.968 & 0.988 & 0.945 & 0.990 \\ \cline{2-12}
    & UNCCD-OOS & 0.997 & 0.972 & 0.983 & 0.977 & 0.968 & 0.981 & 0.927 & 0.983 & 0.887 & 0.982 \\ \cline{2-12}
    & RKCCD-IOS & 0.999 & 0.923 & 0.996 & 0.954 & 0.995 & 0.974 & 0.989 & 0.994 & 0.983 & 0.998 \\ \cline{2-12}
    & UNCCD-IOS & 1.000 & 0.980 & 0.999 & 0.989 & 0.998 & 0.995 & 0.998 & 0.999 & 0.996 & 1.000 \\ \cline{1-12}

    \multirow{4}*{$d=50$} & RKCCD-OOS & 0.994 & 0.989 & 1.000 & 0.999 & 1.000 & 1.000 & 1.000 & 1.000 & 0.998 & 1.000 \\ \cline{2-12}
    & UNCCD-OOS & 1.000 & 0.935 & 1.000 & 0.971 & 1.000 & 0.984 & 0.999 & 0.990 & 0.998 & 0.991 \\ \cline{2-12}
    & RKCCD-IOS & 0.932 & 0.990 & 0.770 & 0.993 & 0.633 & 0.993 & 0.473 & 0.993 & 0.399 & 0.993 \\ \cline{2-12}
    & UNCCD-IOS & 0.987 & 0.994 & 0.982 & 0.998 & 0.999 & 1.000 & 0.999 & 1.000 & 0.998 & 0.995 \\ \cline{1-12}

    \multirow{4}*{$d=100$} & RKCCD-OOS & 0.996 & 0.997 & 1.000 & 1.000 & 1.000 & 1.000 & 1.000 & 1.000 & 1.000 & 1.000 \\ \cline{2-12}
    & UNCCD-OOS & 0.996 & 0.941 & 1.000 & 0.979 & 1.000 & 0.990 & 1.000 & 0.993 & 1.000 & 0.993 \\ \cline{2-12}
    & RKCCD-IOS & 0.944 & 0.992 & 0.789 & 0.994 & 0.686 & 0.994 & 0.548 & 0.994 & 0.479 & 0.994 \\ \cline{2-12}
    & UNCCD-IOS & 0.774 & 0.995 & 0.647 & 0.996 & 0.639 & 0.997 & 0.775 & 0.998 & 0.897 & 0.998 \\ \cline{1-12}
  \end{tabular}}
  \caption{Summary of the TPRs and TNRs of all the CCD-based OSs, with the simulation settings (with Gaussian clusters) elaborated in our previous work \cite{shi2024outlier}.}\label{tab:Gau_General_Results_OS1}
\end{table}


\begin{table}[ht]
  \centering
  \resizebox{\columnwidth}{!}{\begin{tabular}{|c|c|c|c|c|c|c|c|c|c|c|c|}
    \hline
    \multicolumn{2}{|c|}{} & \multicolumn{10}{|c|}{The Size of Datasets} \\ \cline{3-12}

    \multicolumn{2}{|c|}{} & \multicolumn{2}{|c|}{50} & \multicolumn{2}{|c|}{100} & \multicolumn{2}{|c|}{200} & \multicolumn{2}{|c|}{500} & \multicolumn{2}{|c|}{1000} \\ \cline{3-12}

    \multicolumn{2}{|c|}{} & BA & $F_2$-score & BA & $F_2$-score & BA & $F_2$-score & BA & $F_2$-score & BA & $F_2$-score \\ \hline

    \multirow{4}*{$d=2$} & RKCCD-OOS & 0.973 & 0.886 & 0.964 & 0.873 & 0.942 & 0.841 & 0.839 & 0.676 & 0.725 & 0.478 \\ \cline{2-12}
    & UNCCD-OOS & 0.964 & 0.894 & 0.925 & 0.833 & 0.878 & 0.758 & 0.734 & 0.507 & 0.626 & 0.293 \\ \cline{2-12}
    & RKCCD-IOS & 0.984 & 0.930 & 0.986 & 0.926 & 0.987 & 0.917 & 0.987 & 0.912 & 0.987 & 0.912 \\ \cline{2-12}
    & UNCCD-IOS & 0.981 & 0.936 & 0.965	& 0.898 & 0.978 & 0.903 & 0.979 & 0.890 & 0.982 & 0.886 \\ \cline{1-12}

    \multirow{4}*{$d=3$} & RKCCD-OOS & 0.983 & 0.913 & 0.979 & 0.907 & 0.955 & 0.871 & 0.893 & 0.770 & 0.801 & 0.614 \\ \cline{2-12}
    & UNCCD-OOS & 0.984 & 0.915 & 0.971 & 0.893 & 0.949 & 0.862 & 0.881 & 0.755 & 0.793 & 0.606 \\ \cline{2-12}
    & RKCCD-IOS & 0.987 & 0.912 & 0.986 & 0.909 & 0.985 & 0.902 & 0.985 & 0.902 & 0.986 & 0.903 \\ \cline{2-12}
    & UNCCD-IOS & 0.981 & 0.911 & 0.979 & 0.893 & 0.981 & 0.884 & 0.981 & 0.873 & 0.980 & 0.868 \\ \cline{1-12}

    \multirow{4}*{$d=5$} & RKCCD-OOS & 0.988 & 0.924 & 0.978 & 0.914 & 0.962 & 0.891 & 0.906 & 0.799 & 0.830 & 0.672 \\ \cline{2-12}
    & UNCCD-OOS & 0.987 & 0.922 & 0.981 & 0.913 & 0.970 & 0.899 & 0.943 & 0.859 & 0.898 & 0.786 \\ \cline{2-12}
    & RKCCD-IOS & 0.991 & 0.946 & 0.988 & 0.942 & 0.985 & 0.929 & 0.980 & 0.902 & 0.975 & 0.892 \\ \cline{2-12}
    & UNCCD-IOS & 0.987 & 0.925 & 0.985 & 0.907 & 0.985 & 0.897 & 0.985 & 0.900 & 0.986 & 0.903 \\ \cline{1-12}

    \multirow{4}*{$d=10$} & RKCCD-OOS & 0.990 & 0.929 & 0.991 & 0.944 & 0.989 & 0.943 & 0.977 & 0.921 & 0.951 & 0.881 \\ \cline{2-12}
    & UNCCD-OOS & 0.979 & 0.910 & 0.951 & 0.873 & 0.919 & 0.825 & 0.870 & 0.749 & 0.857 & 0.726 \\ \cline{2-12}
    & RKCCD-IOS & 0.986 & 0.903 & 0.989 & 0.925 & 0.991 & 0.940 & 0.989 & 0.935 & 0.981 & 0.910 \\ \cline{2-12}
    & UNCCD-IOS & 0.980 & 0.869 & 0.978 & 0.860 & 0.976 & 0.846 & 0.976 & 0.843 & 0.989 & 0.920 \\ \cline{1-12}

    \multirow{4}*{$d=20$} & RKCCD-OOS & 0.987 & 0.912 & 0.988 & 0.937 & 0.986 & 0.941 & 0.978 & 0.931 & 0.968 & 0.920 \\ \cline{2-12}
    & UNCCD-OOS & 0.985 & 0.902 & 0.980 & 0.907 & 0.975 & 0.908 & 0.955 & 0.883 & 0.935 & 0.848 \\ \cline{2-12}
    & RKCCD-IOS & 0.961 & 0.773 & 0.975 & 0.848 & 0.985 & 0.906 & 0.992 & 0.969 & 0.991 & 0.979 \\ \cline{2-12}
    & UNCCD-IOS & 0.990 & 0.929 & 0.994 & 0.959 & 0.997 & 0.980 & 0.998 & 0.994 & 0.998 & 0.997 \\ \cline{1-12}

    \multirow{4}*{$d=50$} & RKCCD-OOS & 0.992 & 0.955 & 1.000 & 0.996 & 1.000 & 1.000 & 1.000 & 0.999 & 0.999 & 0.998 \\ \cline{2-12}
    & UNCCD-OOS & 0.968 & 0.802 & 0.986 & 0.901 & 0.992 & 0.943 & 0.995 & 0.963 & 0.995 & 0.965 \\ \cline{2-12}
    & RKCCD-IOS & 0.961 & 0.910 & 0.882 & 0.785 & 0.813 & 0.664 & 0.733 & 0.513 & 0.696 & 0.440 \\ \cline{2-12}
    & UNCCD-IOS & 0.991 & 0.967 & 0.990 & 0.978 & 1.000 & 0.999 & 1.000 & 0.999 & 0.997 & 0.980 \\ \cline{1-12}

    \multirow{4}*{$d=100$} & RKCCD-OOS & 0.997 & 0.986 & 1.000 & 1.000 & 1.000 & 1.000 & 1.000 & 1.000 & 1.000 & 1.000 \\ \cline{2-12}
    & UNCCD-OOS & 0.969 & 0.814 & 0.990 & 0.926 & 0.995 & 0.963 & 0.997 & 0.974 & 0.997 & 0.974 \\ \cline{2-12}
    & RKCCD-IOS & 0.968 & 0.926 & 0.892 & 0.805 & 0.840 & 0.715 & 0.771 & 0.588 & 0.737 & 0.521 \\ \cline{2-12}
    & UNCCD-IOS & 0.885 & 0.795 & 0.822 & 0.685 & 0.818 & 0.680 & 0.887 & 0.805 & 0.948 & 0.909 \\ \cline{1-12}
  \end{tabular}}
  \caption{Summary of the BAs and $F_2$-scores of all the CCD-based OSs, with the simulation settings (with Gaussian clusters) elaborated in our previous work \cite{shi2024outlier}.}\label{tab:Gau_General_Results_OS2}
\end{table}


We start with RKCCD-OOS and UNCCD-OOS. 
Both OOSs exhibit similar behavior regarding $n$ and $d$, 
aligning with the patterns observed in simulations with uniform clusters. 
Specifically, when $d\leq5$, 
the two OSs degrade as $n$ increases due to the masking problem, 
raised by the growing intensity of outliers. 
When $d$ is large, the two OOSs achieve promising results, 
with most $F_2$-scores exceeding 0.90. 
Notably, when $d=10$, RKCCD-OOS outperforms UNCCD-OOS (e.g., when $n=1000$ and $d=10$, the $F_2$-scores of two OSSs are 0.881 and 0.726 respectively), 
because UN-CCD may have difficulty distinguishing the regular points near the Gaussian distribution edge and the outliers.

Next, we consider RKCCD-IOS and UNCCD-IOS. 
Both IOSs achieve high $F_2$-scores when $d \leq 20$, 
because both IOSs address the masking problem raised by high-intensity outliers well. 
However, the performance of RKCCD-IOS drops substantially when $d \geq 50 $, 
particularly when $n \geq 500$, 
For instance, when $d=100$ and $n=1000$, the $F_2$-score of RKCCD-IOS is 0.521.
This limitation stems from the same issue encountered when using RK-CCDs for clustering, as detailed in \cite{shi2024outlier}.


Then, we compare the simulation results of the OSs with the U-MCCD and UN-MCCD methods.
(whose performance can be found in our previous work \cite{shi2024outlier})

We first compare RKCCD-OOS and RKCCD-IOS with their prototype, the U-MCCD method. 
The U-MCCD method degrades as $n$ or $d$ increases due to its inability to differentiate outliers from regular points at the edge of Gaussian distributions, 
leading to many false positives. 
Both OSs, generally outperform the U-MCCD method. 
Similar to the simulation settings with uniform clusters, 
RKCCD-IOS outperforms the other two when $d \leq 20$. 
In contrast, RKCCD-OOS is a better choice in high-dimensional data with $F_2$-scores larger than 0.9 even when $d \geq 50$.

Next, we compare UNCCD-OOS, UNCCD-IOS, and the UN-MCCD method,
which show varying performance.
The UN-MCCD method, while superior to U-MCCD, 
exhibits limitations because it constructs a single covering ball for each cluster, 
which is inadequate to cover a Gaussian cluster. 
In contrast, both UNCCD-OOS and UNCCD-IOS demonstrate substantial improvement. 
Although the $F_2$-score of UNCCD-OOS may be as low as 0.293 (e.g., when $d=2$ and $n=1000$) due to the masking problem under lower dimensions, 
and occasionally underperforms compared to the UN-MCCD method, 
UNCCD-IOS consistently outperforms the UN-MCCD method and achieves considerably higher $F_2$-scores across all the simulation settings.

Based on our simulation results,
we observe that RKCCD-OOS and UNCCD-OOS deliver good performance except when $d \leq 5$ and $n$ is large, 
where the masking problem emerges. 
In contrast, RKCCD-IOS and UNCCD-IOS perform much better with lower dimensions but may degrade when $d$ is large because both RK-CCDs and UN-CCDs (especially RK-CCDs) may identify a set of close outliers as valid clusters in high dimensions. 
Compared to the U-MCCD and UN-MCCD methods, 
the CCD-based OSs return fewer false positives, 
achieving much higher $F_2$-scores in general. 
Moreover, they show substantial advantages in higher-dimensional datasets or datasets with Gaussian clusters, 
making them better choices in those cases.

\subsection{Under Random Cluster Process}

\label{sec:Flex_Simul_OS}
In the previous part,  
we have simulated three Neyman-Scott cluster processes (the Mat\'{e}rn, Thomas, and mixed cluster processes) to introduce randomness in both the location and size of clusters, 
The four CCD-based outlier detection methods were compared to several established state-of-the-art approaches, 
including \textbf{Local Outlier Factor} (LOF) \cite{breunig2000lof}, \textbf{Density Based Spatial Clustering of Applications with Noise} (DBSCAN) \cite{ester1996density}, the \textbf{Minimal Spanning Tree} (MST) Method \cite{MST}, \textbf{Outlier Detection using In-degree Number} (ODIN) \cite{hautamaki2004outlier}, and \textbf{isolation Forest} (iForest) \cite{liu2008isolation}. 
Simulation analysis shows LOF achieved the highest $F_2$ scores in over half of the simulation settings, 
and the SUN-MCCD method demonstrated the second-best overall performance while simultaneously providing clustering results. 
Moreover, it is worth noting that the SUN-MCCD method outperforms other cluster-based outlier detection methods by substantial margins, which makes it a decent choice under most simulation settings.

We assess the performance of the four outlying scores by repeating the random cluster process in Tech Report 1. 
For comparison, this analysis included the four CCD-based methods and existing outlier detection methods. 
To expand on the previous evaluation, 
we increase the maximum dimensions from 20 to 100 to evaluate their performance in high-dimensional space.
As noted in Section \ref{sec:Simulation_OS}, 
IOS can be affected by the masking problem. 
This issue is exacerbated as the data size $n$ goes from 50 to 1000 in low-dimensional settings due to the increase in outlier intensity. 
To address this,
we introduce a parameter, $S_{min}$, set to 0.04, 
for the two IOSs (excluding OOSs since they are not cluster-based), 
filtering the clusters with sizes below $S_{min}$. 
We use the same thresholds in Tables \ref{tab:Outlying_Score_NThreshold} and \ref{tab:Outlying_Score_GThreshold} for the Mat\'{e}rn cluster process and the Thomas cluster process, respectively. 
For the mixed cluster process, we set the thresholds as the mean of the values from the two tables. 
For example, when $d=100$, the thresholds of RKCCD-OOS are 13 (Table \ref{tab:Outlying_Score_NThreshold}) and 10 (Table \ref{tab:Outlying_Score_GThreshold}),
resulting in a threshold of 11.5 for the mixed cluster process.

Appropriate parameter selections for the benchmark methods is described as follows:
 
For LOF, the lower and upper bounds of $k$ are 11 and 30, respectively. 
The highest LOF is identified for each point, with a threshold of 1.5. 
In DBSCAN, the outlier percentage is assumed to be known to get an appropriate cutoff value for the $4$-dist, 
serving as a threshold for outliers.
For MST, inconsistent edges are identified using a threshold of 1.2, 
and clusters smaller than 2\% of the dataset are labeled as outliers. 
As for ODIN, we set the input parameters $k$ and $T$ to $n^{0.5}$ and $n^{0.33}$, where $n$ is the data size; 
iForests are constructed with 1000 iTrees, each with a sub-sample size of 256, 
a threshold of 0.55 is applied to capture the outliers.

The simulation results are summarized in Tables \ref{tab:Matern_score1} to \ref{Matern-Thomas_score2}. 
For each simulation setting, we rank the performance of methods by their $F_2$-scores in Table \ref{tab:Complex_Cluster_Ranking2}, and the top 3 are highlighted in bold.
  
\begin{table}[htb]
  \resizebox{\textwidth}{!}{\begin{tabular}{|c|c|c|c|c|c|c|c|c|c|c|c|c|c|c|}
  \hline
  \multirow{2}*{Algorithms} & \multicolumn{2}{|c|}{$d=2$} & \multicolumn{2}{|c|}{$d=3$} & \multicolumn{2}{|c|}{$d=5$} & \multicolumn{2}{|c|}{$d=10$} & \multicolumn{2}{|c|}{$d=20$} & \multicolumn{2}{|c|}{$d=50$} & \multicolumn{2}{|c|}{$d=100$} \\ \cline{2-15}
   & TPR & TNR & TPR & TNR & TPR & TNR & TPR & TNR & TPR & TNR & TPR & TNR & TPR & TNR \\ \hline
  U-MCCDs   & 0.949 & 0.926	& 0.941 & 0.932 & 0.973 & 0.923 & 0.982 & 0.828 & 0.981 & 0.654 & 1.000 & 0.603 & 0.999 & 0.599 \\ \hline
  SU-MCCDs  & 0.969 & 0.954 & 0.970 & 0.940 & 0.971 & 0.945 & 0.982 & 0.849 & 0.979 & 0.678 & 1.000 & 0.603 & 0.999 & 0.599 \\ \hline
  UN-MCCDs  & 0.939 & 0.931 & 0.940 & 0.936 & 0.942 & 0.957 & 0.978 & 0.948 & 0.978 & 0.841 & 0.976 & 0.519 & 0.997 & 0.289 \\ \hline
  SUN-MCCDs & 0.952 & 0.948 & 0.970 & 0.932 & 0.940 & 0.973 & 0.977 & 0.961 & 0.977 & 0.853 & 0.974 & 0.552 & 0.997 & 0.289 \\ \hline
  RKCCD-OOS & 0.898 & 0.924 & 0.891 & 0.945 & 0.816 & 0.955 & 0.750 & 0.993 & 0.539 & 0.991 & 0.314 & 0.907 & 0.311 & 0.905 \\ \hline
  RKCCD-IOS & 0.956 & 0.973 & 0.952 & 0.977 & 0.976 & 0.966 & 0.991 & 0.984 & 0.997 & 0.990 & 0.997 & 1.000 & 0.999 & 1.000 \\ \hline
  UNCCD-OOS & 0.699 & 0.971 & 0.824 & 0.985 & 0.851 & 0.995 & 0.734 & 0.995 & 0.643 & 0.994 & 0.417 & 0.940 & 0.314 & 0.904 \\ \hline
  UNCCD-IOS & 0.949 & 0.953 & 0.955 & 0.970 & 0.958 & 0.979 & 0.985 & 0.978 & 0.988 & 0.984 & 0.998 & 0.999 & 1.000 & 1.000 \\ \hline
  LOF       & 0.999 & 0.962 & 0.999 & 0.962 & 1.000 & 0.927 & 0.999 & 0.866 & 0.999 & 0.842 & 0.998 & 0.815 & 0.996 & 0.821 \\ \hline
  DBSCAN    & 0.891 & 0.988 & 0.789 & 0.996 & 0.768 & 1.000 & 0.771 & 1.000 & 0.750 & 1.000 & 0.732 & 0.999 & 0.722 & 0.999 \\ \hline
  MST       & 0.659 & 0.661 & 0.558 & 0.875 & 0.623 & 0.881 & 0.713 & 0.855 & 0.757 & 0.802 & 0.751 & 0.790 & 0.661 & 0.924 \\ \hline
  ODIN      & 0.912 & 0.937 & 0.918 & 0.977 & 0.905 & 0.988 & 0.898 & 0.991 & 0.870 & 0.993 & 0.843 & 0.993 & 0.822 & 0.993 \\ \hline
  iForest   & 0.855 & 0.904 & 0.756 & 0.946 & 0.800 & 0.967 & 0.915 & 0.974 & 0.982 & 0.972 & 0.999 & 0.964 & 1.000 & 0.951 \\ \hline
\end{tabular}}
  \caption{The TPRs and TNRs of selected outlier detection methods and OSs under a Mat\'{e}rn cluster process (the simulation setting \uppercase\expandafter{\romannumeral1} in our previous work \cite{shi2024outlier}).}\label{tab:Matern_score1}
\end{table}

\begin{table}[htb]
  \resizebox{\textwidth}{!}{\begin{tabular}{|c|c|c|c|c|c|c|c|c|c|c|c|c|c|c|}
  \hline
  \multirow{2}*{Algorithms} & \multicolumn{2}{|c|}{$d=2$} & \multicolumn{2}{|c|}{$d=3$} & \multicolumn{2}{|c|}{$d=5$} & \multicolumn{2}{|c|}{$d=10$} & \multicolumn{2}{|c|}{$d=20$} & \multicolumn{2}{|c|}{$d=50$} & \multicolumn{2}{|c|}{$d=100$} \\ \cline{2-15}
   & BA & $F_2$-score & BA & $F_2$-score & BA & $F_2$-score & BA & $F_2$-score & BA & $F_2$-score & BA & $F_2$-score & BA & $F_2$-score \\
  \hline
  U-MCCDs   & 0.938 & 0.732 & 0.937 & 0.824 & 0.948 & 0.853 & 0.905 & 0.747 & 0.818 & 0.595 & 0.802 & 0.579 & 0.799 & 0.576 \\ \hline
  SU-MCCDs  & 0.962 & 0.822 & 0.955 & 0.863 & 0.958 & 0.886 & 0.916 & 0.886 & 0.829 & 0.610 & 0.802 & 0.580 & 0.799 & 0.576 \\ \hline
  UN-MCCDs  & 0.935 & 0.730 & 0.938 & 0.833 & 0.950 & 0.875 & 0.963 & 0.892 & 0.910 & 0.755 & 0.748 & 0.515 & 0.643 & 0.431 \\ \hline
  SUN-MCCDs & 0.950 & 0.787 & 0.951 & 0.851 & 0.957 & 0.902 & 0.969 & 0.912 & 0.915 & 0.768 & 0.763 & 0.531 & 0.643 & 0.431 \\ \hline
  RKCCD-OOS & 0.911 & 0.675 & 0.918 & 0.793 & 0.886 & 0.764 & 0.872 & 0.771 & 0.765 & 0.573 & 0.611 & 0.262 & 0.608 & 0.259 \\ \hline
  RKCCD-IOS & 0.965 & 0.840 & 0.965 & 0.898 & 0.971 & 0.909 & 0.988 & 0.959 & 0.994 & 0.974 & 0.999 & 0.997 & 1.000 & 0.999 \\ \hline
  UNCCD-OOS & 0.835 & 0.630 & 0.905 & 0.815 & 0.923 & 0.865 & 0.865 & 0.762 & 0.819 & 0.677 & 0.679 & 0.407 & 0.609 & 0.261 \\ \hline
  UNCCD-IOS & 0.951 & 0.770 & 0.963 & 0.884 & 0.969 & 0.917 & 0.982 & 0.938 & 0.986 & 0.954 & 0.999 & 0.997 & 1.000 & 1.000 \\ \hline
  LOF       & 0.981 & 0.866 & 0.981 & 0.926 & 0.964 & 0.884 & 0.933 & 0.802 & 0.921 & 0.774 & 0.907 & 0.746 & 0.909 & 0.749 \\ \hline
  DBSCAN    & 0.940 & 0.827 & 0.893 & 0.794 & 0.884 & 0.786 & 0.886 & 0.789 & 0.875 & 0.767 & 0.866 & 0.750 & 0.861 & 0.739 \\ \hline
  MST       & 0.660 & 0.283 & 0.717 & 0.450 & 0.752 & 0.525 & 0.784 & 0.556 & 0.780 & 0.536 & 0.771 & 0.524 & 0.793 & 0.578 \\ \hline
  ODIN      & 0.925 & 0.783 & 0.948 & 0.882 & 0.947 & 0.901 & 0.945 & 0.901 & 0.932 & 0.879 & 0.918 & 0.855 & 0.908 & 0.837 \\ \hline
  iForest   & 0.880 & 0.615 & 0.851 & 0.691 & 0.884 & 0.775 & 0.945 & 0.877 & 0.977 & 0.925 & 0.982 & 0.923 & 0.976 & 0.901 \\ \hline
\end{tabular}}
  \caption{The BAs and $F_2$-scores of selected outlier detection methods and OSs under a Mat\'{e}rn cluster process (the simulation setting \uppercase\expandafter{\romannumeral1} in our previous work \cite{shi2024outlier}).}\label{Matern_score2}
\end{table}

\begin{table}[htb]
  \resizebox{\textwidth}{!}{\begin{tabular}{|c|c|c|c|c|c|c|c|c|c|c|c|c|c|c|}
  \hline
  \multirow{2}*{Algorithms} & \multicolumn{2}{|c|}{$d=2$} & \multicolumn{2}{|c|}{$d=3$} & \multicolumn{2}{|c|}{$d=5$} & \multicolumn{2}{|c|}{$d=10$} & \multicolumn{2}{|c|}{$d=20$} & \multicolumn{2}{|c|}{$d=50$} & \multicolumn{2}{|c|}{$d=100$} \\ \cline{2-15}
   & TPR & TNR & TPR & TNR & TPR & TNR & TPR & TNR & TPR & TNR & TPR & TNR & TPR & TNR \\\hline
  U-MCCDs   & 0.924 & 0.907 & 0.966 & 0.852 & 0.987 & 0.772 & 0.976 & 0.669 & 0.974 & 0.497 & 0.980 & 0.574 & 0.934 & 0.600 \\ \hline
  SU-MCCDs  & 0.880 & 0.959 & 0.942 & 0.922 & 0.983 & 0.849 & 0.976 & 0.734 & 0.973 & 0.534 & 0.980 & 0.574 & 0.937 & 0.595 \\ \hline
  UN-MCCDs  & 0.824 & 0.963 & 0.918 & 0.941 & 0.970 & 0.922 & 0.989 & 0.889 & 0.979 & 0.744 & 0.981 & 0.365 & 0.999 & 0.402 \\ \hline
  SUN-MCCDs & 0.632 & 0.978 & 0.853 & 0.951 & 0.964 & 0.900 & 0.973 & 0.836 & 0.971 & 0.745 & 0.988 & 0.207 & 1.000 & 0.517 \\ \hline
  RKCCD-OOS & 0.752 & 0.925 & 0.721 & 0.945 & 0.792 & 0.942 & 0.854 & 0.940 & 0.717 & 0.922 & 0.540 & 0.921 & 0.547 & 0.943 \\ \hline
  RKCCD-IOS & 0.891 & 0.951 & 0.918 & 0.936 & 0.941 & 0.919 & 0.943 & 0.898 & 0.809 & 0.997 & 0.912 & 0.999 & 0.924 & 0.997 \\ \hline
  UNCCD-OOS & 0.544 & 0.956 & 0.659 & 0.961 & 0.849 & 0.953 & 0.805 & 0.942 & 0.766 & 0.942 & 0.570 & 0.898 & 0.548 & 0.943 \\ \hline
  UNCCD-IOS & 0.868 & 0.953 & 0.897 & 0.949 & 0.929 & 0.941 & 0.899 & 0.963 & 0.914 & 0.917 & 0.927 & 0.980 & 0.924 & 0.981 \\ \hline
  LOF       & 0.979 & 0.943 & 0.960 & 0.960 & 0.967 & 0.961 & 0.997 & 0.921 & 0.996 & 0.862 & 0.988 & 0.839 & 0.982 & 0.835 \\ \hline
  DBSCAN    & 0.824 & 0.990 & 0.684 & 0.998 & 0.728 & 0.999 & 0.726 & 0.999 & 0.707 & 0.999 & 0.646 & 0.996 & 0.571 & 0.993 \\ \hline
  MST       & 0.405 & 0.866 & 0.336 & 0.947 & 0.587 & 0.917 & 0.634 & 0.911 & 0.633 & 0.967 & 0.526 & 0.973 & 0.555 & 0.966 \\ \hline
  ODIN      & 0.891 & 0.930 & 0.903 & 0.917 & 0.916 & 0.907 & 0.899 & 0.895 & 0.859 & 0.879 & 0.822 & 0.834 & 0.812 & 0.791 \\ \hline
  iForest   & 0.857 & 0.892 & 0.708 & 0.938 & 0.644 & 0.961 & 0.716 & 0.975 & 0.789 & 0.972 & 0.928 & 0.953 & 0.973 & 0.938 \\ \hline
\end{tabular}}
  \caption{The TPRs and TNRs of selected outlier detection methods and OSs under a Thomas cluster process (the simulation setting \uppercase\expandafter{\romannumeral2} in our previous work \cite{shi2024outlier}).}\label{Thomas_score1}
\end{table}

\begin{table}[htb]
  \resizebox{\textwidth}{!}{\begin{tabular}{|c|c|c|c|c|c|c|c|c|c|c|c|c|c|c|}
  \hline
  \multirow{2}*{Algorithms} & \multicolumn{2}{|c|}{$d=2$} & \multicolumn{2}{|c|}{$d=3$} & \multicolumn{2}{|c|}{$d=5$} & \multicolumn{2}{|c|}{$d=10$} & \multicolumn{2}{|c|}{$d=20$} & \multicolumn{2}{|c|}{$d=50$} & \multicolumn{2}{|c|}{$d=100$} \\ \cline{2-15}
   & BA & $F_2$-score & BA & $F_2$-score & BA & $F_2$-score & BA & $F_2$-score & BA & $F_2$-score & BA & $F_2$-score & BA & $F_2$-score \\
  \hline
  U-MCCDs   & 0.916 & 0.611 & 0.909 & 0.706 & 0.880 & 0.682 & 0.823 & 0.603 & 0.736 & 0.509 & 0.777 & 0.564 & 0.767 & 0.573 \\ \hline
  SU-MCCDs  & 0.920 & 0.711 & 0.932 & 0.794 & 0.916 & 0.763 & 0.855 & 0.652 & 0.754 & 0.526 & 0.777 & 0.564 & 0.766 & 0.557 \\ \hline
  UN-MCCDs  & 0.904 & 0.639 & 0.920 & 0.751 & 0.920 & 0.756 & 0.906 & 0.743 & 0.820 & 0.601 & 0.666 & 0.455 & 0.697 & 0.520 \\ \hline
  SUN-MCCDs & 0.894 & 0.687 & 0.930 & 0.806 & 0.946 & 0.845 & 0.939 & 0.822 & 0.862 & 0.664 & 0.673 & 0.465 & 0.701 & 0.523 \\ \hline
  RKCCD-OOS & 0.839 & 0.518 & 0.833 & 0.646 & 0.867 & 0.724 & 0.897 & 0.775 & 0.820 & 0.648 & 0.731 & 0.506 & 0.745 & 0.563 \\ \hline
  RKCCD-IOS & 0.921 & 0.678 & 0.927 & 0.782 & 0.930 & 0.804 & 0.921 & 0.783 & 0.903 & 0.831 & 0.956 & 0.925 & 0.961 & 0.931 \\ \hline
  UNCCD-OOS & 0.750 & 0.437 & 0.810 & 0.624 & 0.901 & 0.785 & 0.874 & 0.741 & 0.854 & 0.711 & 0.734 & 0.509 & 0.746 & 0.538 \\ \hline
  UNCCD-IOS & 0.911 & 0.667 & 0.923 & 0.789 & 0.935 & 0.826 & 0.931 & 0.844 & 0.916 & 0.786 & 0.954 & 0.905 & 0.953 & 0.915 \\ \hline
  LOF       & 0.961 & 0.741 & 0.960 & 0.877 & 0.964 & 0.908 & 0.959 & 0.876 & 0.929 & 0.802 & 0.914 & 0.773 & 0.909 & 0.778 \\ \hline
  DBSCAN    & 0.907 & 0.740 & 0.841 & 0.708 & 0.864 & 0.751 & 0.863 & 0.744 & 0.853 & 0.726 & 0.821 & 0.662 & 0.782 & 0.585 \\ \hline
  MST       & 0.636 & 0.233 & 0.642 & 0.320 & 0.752 & 0.535 & 0.773 & 0.561 & 0.800 & 0.623 & 0.750 & 0.530 & 0.761 & 0.548 \\ \hline
  ODIN      & 0.911 & 0.634 & 0.910 & 0.749 & 0.912 & 0.778 & 0.897 & 0.759 & 0.869 & 0.713 & 0.828 & 0.635 & 0.802 & 0.595 \\ \hline
  iForest   & 0.875 & 0.545 & 0.823 & 0.632 & 0.803 & 0.633 & 0.846 & 0.717 & 0.881 & 0.774 & 0.941 & 0.850 & 0.956 & 0.860 \\ \hline
\end{tabular}}
  \caption{The BAs and $F_2$-scores of selected outlier detection methods and OSs under Thomas cluster process (the simulation setting \uppercase\expandafter{\romannumeral2} in our previous work \cite{shi2024outlier}).}\label{Thomas_score2}
\end{table}

\begin{table}[htb]
  \resizebox{\textwidth}{!}{\begin{tabular}{|c|c|c|c|c|c|c|c|c|c|c|c|c|c|c|}
  \hline
  \multirow{2}*{Algorithms} & \multicolumn{2}{|c|}{$d=2$} & \multicolumn{2}{|c|}{$d=3$} & \multicolumn{2}{|c|}{$d=5$} & \multicolumn{2}{|c|}{$d=10$} & \multicolumn{2}{|c|}{$d=20$} & \multicolumn{2}{|c|}{$d=50$} & \multicolumn{2}{|c|}{$d=100$} \\ \cline{2-15}
   & TPR & TNR & TPR & TNR & TPR & TNR & TPR & TNR & TPR & TNR & TPR & TNR & TPR & TNR \\
  \hline
  U-MCCDs   & 0.942 & 0.907 & 0.953 & 0.884 & 0.978 & 0.856 & 0.975 & 0.745 & 0.980 & 0.588 & 0.987 & 0.568 & 0.982 & 0.592 \\ \hline
  SU-MCCDs  & 0.925 & 0.952 & 0.955 & 0.921 & 0.975 & 0.900 & 0.974 & 0.775 & 0.974 & 0.617 & 0.987 & 0.568 & 0.982 & 0.592 \\ \hline
  UN-MCCDs  & 0.910 & 0.926 & 0.927 & 0.912 & 0.952 & 0.914 & 0.984 & 0.883 & 0.975 & 0.757 & 0.980 & 0.541 & 0.999 & 0.429 \\ \hline
  SUN-MCCDs & 0.891 & 0.952 & 0.939 & 0.932 & 0.946 & 0.950 & 0.983 & 0.915 & 0.974 & 0.779 & 0.978 & 0.593 & 0.999 & 0.429 \\ \hline
  RKCCD-OOS & 0.752 & 0.946 & 0.753 & 0.962 & 0.747 & 0.977 & 0.812 & 0.972 & 0.630 & 0.964 & 0.473 & 0.792 & 0.367 & 0.808 \\ \hline
  RKCCD-IOS & 0.929 & 0.950 & 0.922 & 0.959 & 0.887 & 0.949 & 0.868 & 0.937 & 0.969 & 0.880 & 0.997 & 0.950 & 1.000 & 0.994 \\ \hline
  UNCCD-OOS & 0.666 & 0.956 & 0.761 & 0.966 & 0.848 & 0.973 & 0.781 & 0.966 & 0.718 & 0.966 & 0.558 & 0.816 & 0.421 & 0.744 \\ \hline
  UNCCD-IOS & 0.949 & 0.952 & 0.955 & 0.970 & 0.958 & 0.979 & 0.985 & 0.977 & 0.988 & 0.983 & 0.998 & 0.999 & 1.000 & 1.000 \\ \hline
  LOF       & 0.990 & 0.948 & 0.984 & 0.957 & 0.984 & 0.942 & 0.998 & 0.893 & 0.993 & 0.857 & 0.980 & 0.894 & 0.998 & 0.953 \\ \hline
  DBSCAN    & 0.849 & 0.988 & 0.789 & 0.996 & 0.749 & 0.998 & 0.746 & 0.998 & 0.725 & 0.997 & 0.611 & 0.993 & 0.673 & 0.997 \\ \hline
  MST       & 0.443 & 0.859 & 0.382 & 0.950 & 0.576 & 0.934 & 0.593 & 0.938 & 0.562 & 0.983 & 0.376 & 0.984 & 0.221 & 0.998 \\ \hline
  ODIN      & 0.899 & 0.943 & 0.906 & 0.944 & 0.911 & 0.952 & 0.885 & 0.956 & 0.827 & 0.968 & 0.718 & 0.977 & 0.743 & 0.989 \\ \hline
  iForest   & 0.851 & 0.898 & 0.730 & 0.941 & 0.708 & 0.960 & 0.837 & 0.963 & 0.955 & 0.943 & 0.999 & 0.929 & 1.000 & 0.970 \\ \hline
\end{tabular}}
  \caption{The TPRs and TNRs of selected outlier detection methods and OSs under a mixed cluster process (the simulation setting \uppercase\expandafter{\romannumeral3} in our previous work \cite{shi2024outlier}).}\label{Matern-Thomas_score1}
\end{table}

\begin{table}[htb]
  \resizebox{\textwidth}{!}{\begin{tabular}{|c|c|c|c|c|c|c|c|c|c|c|c|c|c|c|}
  \hline
  \multirow{2}*{Algorithms} & \multicolumn{2}{|c|}{$d=2$} & \multicolumn{2}{|c|}{$d=3$} & \multicolumn{2}{|c|}{$d=5$} & \multicolumn{2}{|c|}{$d=10$} & \multicolumn{2}{|c|}{$d=20$} & \multicolumn{2}{|c|}{$d=50$} & \multicolumn{2}{|c|}{$d=100$} \\ \cline{2-15}
   & BA & $F_2$-score & BA & $F_2$-score & BA & $F_2$-score & BA & $F_2$-score & BA & $F_2$-score & BA & $F_2$-score & BA & $F_2$-score \\
  \hline
  U-MCCDs   & 0.925 & 0.660 & 0.919 & 0.752 & 0.917 & 0.763 & 0.860 & 0.658 & 0.784 & 0.550 & 0.778 & 0.556 & 0.787 & 0.566 \\ \hline
  SU-MCCDs  & 0.939 & 0.756 & 0.938 & 0.813 & 0.938 & 0.819 & 0.875 & 0.685 & 0.796 & 0.582 & 0.778 & 0.556 & 0.787 & 0.566 \\ \hline
  UN-MCCDs  & 0.918 & 0.678 & 0.920 & 0.769 & 0.933 & 0.815 & 0.934 & 0.806 & 0.866 & 0.663 & 0.761 & 0.535 & 0.714 & 0.496 \\ \hline
  SUN-MCCDs & 0.922 & 0.736 & 0.936 & 0.816 & 0.948 & 0.866 & 0.949 & 0.850 & 0.877 & 0.682 & 0.786 & 0.562 & 0.714 & 0.496 \\ \hline
  RKCCD-OOS & 0.849 & 0.592 & 0.858 & 0.707 & 0.862 & 0.739 & 0.892 & 0.790 & 0.797 & 0.620 & 0.633 & 0.326 & 0.588 & 0.242 \\ \hline
  RKCCD-IOS & 0.940 & 0.729 & 0.941 & 0.833 & 0.918 & 0.804 & 0.903 & 0.767 & 0.925 & 0.770 & 0.974 & 0.913 & 0.997 & 0.989 \\ \hline
  UNCCD-OOS & 0.811 & 0.548 & 0.864 & 0.721 & 0.911 & 0.819 & 0.874 & 0.755 & 0.842 & 0.701 & 0.687 & 0.433 & 0.583 & 0.252 \\ \hline
  UNCCD-IOS & 0.951 & 0.773 & 0.963 & 0.884 & 0.969 & 0.917 & 0.981 & 0.938 & 0.986 & 0.952 & 0.999 & 0.997 & 1.000 & 1.000 \\ \hline
  LOF       & 0.969 & 0.794 & 0.971 & 0.899 & 0.963 & 0.889 & 0.946 & 0.835 & 0.925 & 0.785 & 0.937 & 0.829 & 0.976 & 0.921 \\ \hline
  DBSCAN    & 0.919 & 0.776 & 0.893 & 0.794 & 0.874 & 0.764 & 0.872 & 0.762 & 0.861 & 0.736 & 0.802 & 0.617 & 0.835 & 0.685 \\ \hline
  MST       & 0.651 & 0.263 & 0.666 & 0.365 & 0.755 & 0.535 & 0.766 & 0.552 & 0.773 & 0.573 & 0.680 & 0.395 & 0.610 & 0.243 \\ \hline
  ODIN      & 0.921 & 0.699 & 0.925 & 0.804 & 0.932 & 0.842 & 0.921 & 0.832 & 0.898 & 0.802 & 0.848 & 0.721 & 0.866 & 0.760 \\ \hline
  iForest   & 0.875 & 0.580 & 0.836 & 0.659 & 0.834 & 0.685 & 0.900 & 0.798 & 0.949 & 0.854 & 0.964 & 0.876 & 0.985 & 0.935 \\ \hline
\end{tabular}}
  \caption{The BAs and $F_2$-scores of selected outlier detection methods and OSs under a mixed cluster process (the simulation setting \uppercase\expandafter{\romannumeral3} in our previous work \cite{shi2024outlier}).}\label{Matern-Thomas_score2}
\end{table}

\begin{table}[htb]
  \resizebox{\textwidth}{!}{\begin{tabular}{|c|c|c|c|c|c|c|c|c|c|c|c|c|c|c|c|c|c|c|c|c|c|}
  \hline
  & \multicolumn{7}{|c|}{Mat\'{e}rn} & \multicolumn{7}{|c|}{Thomas} & \multicolumn{7}{|c|}{Mixed} \\ \cline{1-22}
  $d$ & 2 & 3 & 5 & 10 & 20 & 50 & 100 & 2 & 3 & 5 & 10 & 20 & 50 & 100 & 2 & 3 & 5 & 10 & 20 & 50 & 100 \\ \hline \hline
  U-MCCDs   & 8 & 8 & 9	& 12 & 11 & 8 & 8 & 9 & 9 & 11 & 12 & 13 & 7 & 7 & 9 & 9 & 10 & 12 & 13 & 8 & 7 \\ \hline
  SU-MCCDs  & 4 & 5 & 5 & 6 & 10 & 7 & 8 & \textbf{3} & \textbf{3} & 7 & 11 & 12 & 7 & 9 & 4 & 5 & 5 & 11 & 11 & 8 & 7 \\ \hline
  UN-MCCDs  & 9 & 7 & 7 & 5 & 8 & 11 & 10 & 7 & 6 & 8 & 8 & 11 & 13 & 13 & 8 & 8 & 7 & 5 & 9 & 10 & 9 \\ \hline
  SUN-MCCDs & 5 & 6 & \textbf{3} & \textbf{3} & 6 & 9 & 10 & 4 & \textbf{2} & \textbf{2} & \textbf{3} & 8 & 12 & 12 & 5 & 4 & \textbf{3} & \textbf{2} & 8 & 7 & 9 \\ \hline
  RKCCD-OOS & 10 & 11 & 12 & 11 & 12 & 13 & 13 & 11 & 10 & 10 & 5 & 9 & 11 & 8 & 10 & 11 & 11 & 7 & 10 & 13 & 13 \\ \hline
  RKCCD-IOS & \textbf{2} & \textbf{2} & \textbf{2} & \textbf{1} & \textbf{1} & \textbf{1} & \textbf{2} & 5 & 5 & 4 & 4 & \textbf{1} & \textbf{1} & \textbf{1} & 6 & \textbf{3} & 8 & 8 & 5 & \textbf{2} & \textbf{2} \\ \hline
  UNCCD-OOS & 11 & 9 & 8 & 10 & 9 & 12 & 12 & 12 & 12 & 5 & 9 & 7 & 10 & 11 & 12 & 10 & 5 & 10 & 7 & 11 & 11 \\ \hline
  UNCCD-IOS & 7 & \textbf{3} & \textbf{1} & \textbf{2} & \textbf{2} & \textbf{1} & \textbf{1} & 6 & 4 & \textbf{3} & \textbf{2} & \textbf{3} & \textbf{2} & \textbf{2} & \textbf{3} & \textbf{2} & \textbf{1} & \textbf{1} & \textbf{1} & \textbf{1} & \textbf{1} \\ \hline
  LOF       & \textbf{1} & \textbf{1} & 6 & 8 & 5 & 6 & 5 & \textbf{1} & \textbf{1} & \textbf{1} & \textbf{1} & \textbf{2} & 4 & 4 & \textbf{1} & \textbf{1} & \textbf{2} & \textbf{3} & 4 & 4 & 4 \\ \hline
  DBSCAN    & \textbf{3} & 10 & 10 & 9 & 7 & 5 & 6 & \textbf{2} & 8 & 9 & 7 & 5 & 5 & 6 & \textbf{2} & 7 & 9 & 9 & 6 & 6 & 6 \\ \hline
  MST       & 13 & 13 & 13 & 13 & 13 & 10 & 7 & 13 & 13 & 13 & 13 & 10 & 9 & 10 & 13 & 13 & 13 & 13 & 12 & 12 & 12 \\ \hline
  ODIN      & 6 & 4 & 4 & 4 & 4 & 4 & 4 & 8 & 7 & 6 & 6 & 6 & 6 & 5 & 7 & 6 & 4 & 4 & \textbf{3} & 5 & 5 \\ \hline
  iForest   & 12 & 12 & 11 & 7 & \textbf{3} & \textbf{3} & \textbf{3} & 10 & 11 & 12 & 10 & 4 & \textbf{3} & \textbf{3} & 11 & 12 & 12 & 6 & \textbf{2} & \textbf{3} & \textbf{3} \\ \hline
\end{tabular}}
  \caption{The rankings (by $F_2$-scores) of all the methods under each simulation setting of this section, top 3 are highlighted in bold. UNCCD-IOS delivers the best overall performance compared to all the other methods.}\label{tab:Complex_Cluster_Ranking2}
\end{table}

Considering the two OOSs, while achieving high TNRs, 
both deliver lower TPRs and $F_2$-scores, 
ranking around $10^{th}$ place out of 13 methods under most simulation settings. 
For instance, the TPRs of RKCCD-OOS are 0.752, 0.751, 0.792, 0.854, 0.717, 0.540, and 0.547 under the Thomas cluster process, 
substantially lower than most other methods. 
The reason stems from the limitation of OOS: 
OOS measures the outlyingness of a point by comparing its vicinity density to its outbound neighbors. 
This approach is not robust to the masking problem. 
Moreover, the masking problem is more severe under the simulations with the three Neyman-Scott cluster processes compared to those with distinct clusters, 
as the random cluster processes are more likely to generate a group of close outlying points.

In contrast to OOS, 
both IOSs achieve superior performance, 
ranking highest among the 13 outlier detection methods. 
While the SUN-MCCD method previously showed promising results in our previous work \cite{shi2024outlier},
outperforming other clustering-based approaches, 
LOF achieved higher $F_2$-scores under most simulation settings. 
Fortunately, the two IOSs demonstrate substantial improvement over the SUN-MCCD method. 
For example, in the mixed point process, 
the $F_2$-scores of UNCCD-IOS are 0.773, 0.884, 0.917, 0.938, 0.952, 0.997, and 1.000, 
substantially higher than those achieved by the SUN-MCCD method (0.736, 0.816, 0.866, 0.850, 0.682, 0.562, and 0.496), 
particularly when $d\geq20$. 
Furthermore, considering the ranking in Table \ref{tab:Complex_Cluster_Ranking2}, 
UNCCD-IOS outperforms LOF under most simulation settings. 
For example, under the Mat\'{e}rn cluster process, 
the $F_2$-scores of UNCCD-IOS are higher than LOF when $d\geq5$. 
Consider RKCCD-IOS, although it yields worse overall performance compared to UNCCD-IOS (because UN-CCDs achieve better clustering results than RK-CCDs), 
it is still comparable to or better than LOF, 
making it the second-best method here.

\section{Real Data Examples}
\label{sec:Real-Data_OS}

In this section, we evaluate the performance of all four CCD-based OSs using real-life datasets, 
and compare them with established methods. 
We also want to investigate if the four OSs outperform previously CCD-based methods proposed in Tech Report 1. 
The datasets, 
obtained from \textbf{Outlier Detection Datasets (ODDS)} \cite{Rayana2016} and \textbf{ELKI Outlier Datasets} \cite{DBLP:journals/corr/abs-1902-03616}, 
are generally more complex than the artificial datasets used in Sections \ref{sec:Simulation_OS}. 
Prior to outlier detection, 
we preprocess the datasets by normalizing all the features. 
Traditional normalization, 
subtracting the sample mean and dividing by the sample standard deviation,  
is not robust to outliers with extreme values \cite{maronna2019robust}. 
To address it, 
we employ a robust alternative way with mean and standard deviation replaced by the median (Med) and the \textbf{Normalized Median Absolute Deviation about the median} (MADN), respectively. 

The details of each dataset are summarized below.

\begin{itemize} \label{tab:Real_Data_Des_OS}
  \item [] \textbf{Brief descriptions of each real-life dataset.}
  \item \textbf{hepatitis}: A dataset contains patients suffering from hepatitis that have died (outliers) or survived (inliers).
  \item \textbf{Lymphography (Lymph)}: This dataset represents patients divided into 4 classes according to radiological examination results. Two classes are represented by only 6 instances and thus considered as outliers.
  \item \textbf{glass}: This dataset consists of 6 types of glass, and the $6^{th}$ type is a minority class, thus marked as outliers, while all other points are inliers.
  \item \textbf{WBC}: This dataset consists of examples of different cancer types, benign (inliers) or malignant (outliers).
  \item \textbf{vertebral}: A dataset with six bio-mechanical features, which are used to classify orthopedic patients either as normal (inliers) or abnormal (outliers).
  \item \textbf{stamps}: A dataset with each observation representing forged (photocopied or scanned+printed) stamps (outliers) or genuine (ink) stamps (inlier). The features are based on the color and printing properties of the stamps.
  \item \textbf{WDBC}: This dataset describes nuclear characteristics for breast cancer diagnosis. We consider examples of benign cancer as inliers and malignant cancer as outliers.
  \item \textbf{vowels}: Four male speakers (classes) uttered two Japanese vowels successively; class (speaker) 1 is used as an outlier. The other speakers (classes) are considered inliers.
  \item \textbf{Thyroid}: This dataset is to determine whether a patient referred to the clinic is hypothyroid, which consists of three classes: normal (not hypothyroid), hyperfunction and subnormal functioning. The hyperfunction class is treated as an outlier class, and the other two classes are inliers.
  \item \textbf{wilt}: This dataset differentiates diseased trees (outliers) from other land covers (inliers).
\end{itemize}

\begin{table}[htb]
  \center
  \footnotesize{\begin{tabular}{|c|c|c|c|}
  \hline
  & $n$ & $d$ & \# of outliers \\ \hline
  hepatitis & 74   & 19 & 7 (9.5\%)   \\ \hline
  lymph     & 148  & 18 & 6 (4.1\%)   \\ \hline
  glass     & 214  & 9  & 10 (4.5\%)  \\ \hline
  WBC       & 223  & 9  & 10 (4.5\%)  \\ \hline
  vertebral & 240  & 6  & 30 (12.5\%) \\ \hline
  stamps    & 340  & 9  & 31 (9.1\%)  \\ \hline
  WDBC      & 367  & 30 & 10 (2.72\%) \\ \hline
  vowels    & 1456 & 12 & 50 (3.4\%)  \\ \hline
  thyroid   & 3772 & 6  & 93 (2.5\%)  \\ \hline
  wilt      & 4735 & 5  & 257 (5.4\%) \\ \hline
\end{tabular}}
 \caption{The size ($n$), dimensionality ($d$), and contamination level of each real-life dataset.}\label{tab:Real_Data_OS}
\end{table}

Parameter selections for the benchmark methods remain the same as in Section \ref{sec:Flex_Simul_OS}.
The four OSs use a threshold of 2, 
and the parameter $S_{min}$ for the two IOSs is set to 0. 
We record TPRs, TNRs, BAs, and $F_2$-scores in Tables \ref{Real_Data_Result_OS1} and \ref{Real_Data_Result_OS2}.

\begin{table}[htb]
  \resizebox{\textwidth}{!}{\begin{tabular}{|c|c|c|c|c|c|c|c|c|c|c|c|c|c|c|c|c|c|c|c|c|}
  \hline
  \multirow{2}*{} & \multicolumn{2}{|c|}{hepatitis} & \multicolumn{2}{|c|}{lymph} & \multicolumn{2}{|c|}{glass} & \multicolumn{2}{|c|}{WBC}  & \multicolumn{2}{|c|}{vertebral} & \multicolumn{2}{|c|}{stamps} & \multicolumn{2}{|c|}{WDBC} & \multicolumn{2}{|c|}{vowels} & \multicolumn{2}{|c|}{thyroid} & \multicolumn{2}{|c|}{wilt} \\ \cline{2-21}
  & TPR & TNR & TPR & TNR & TPR & TNR & TPR & TNR & TPR & TNR & TPR & TNR & TPR & TNR & TPR & TNR & TPR & TNR & TPR & TNR\\ \hline
  U-MCCDs   & 0.286 & 0.881 & 0.333 & 0.866 & 1.000 & 0.363 & 0.500 & 0.511 & 0.467 & 0.643 & 0.065 & 0.958 & 0.500 & 0.714 & 1.000 & 0.327 & 0.484 & 0.577 & 0.763 & 0.630 \\ \hline
  SU-MCCDs  & 0.286 & 0.925 & 0.333 & 0.866 & 1.000 & 0.363 & 0.500 & 0.610 & 0.200 & 0.576 & 0.516	& 0.883 & 0.400 & 0.723 & 1.000 & 0.373 & 0.484 & 0.718 & 0.300 & 0.785 \\ \hline
  UN-MCCDs  & 0.714 & 0.657 & 0.333 & 0.810 & 0.222 & 0.765 & 0.500 & 0.474 & 0.033 & 0.914 & 0.484 & 0.812 & 1.000 & 0.182 & 1.000 & 0.541 & 0.484 & 0.616 & 0.140 & 0.897 \\ \hline
  SUN-MCCDs & 0.714 & 0.657 & 0.333 & 0.711 & 1.000 & 0.540 & 0.500 & 0.540 & 0.100 & 0.928 & 0.516	& 0.884 & 1.000 & 0.325 & 0.978 & 0.676 & 0.484 & 0.785 & 0.366 & 0.745 \\ \hline
  RKCCD-OOS & 0.000 & 0.866 & 0.500 & 0.725 & 0.222 & 0.784 & 0.200 & 0.850 & 0.133 & 0.905 & 0.258 & 0.864 & 0.700 & 0.807 & 0.326 & 0.900 & 0.280 & 0.885 & 0.105 & 0.912 \\ \hline
  RKCCD-IOS & 0.142 & 0.925 & 0.833 & 0.789 & 0.333 & 0.765 & 1.000 & 0.779 & 0.067 & 0.843 & 0.226 & 0.838 & 0.700 & 0.913 & 0.783 & 0.898 & 0.828 & 0.842 & 0.304 & 0.810 \\ \hline
  UNCCD-OOS & 0.289 & 0.866 & 0.833 & 0.697 & 0.222 & 0.828 & 0.200 & 0.897 & 0.100 & 0.905 & 0.194 & 0.887 & 0.700 & 0.874 & 0.413 & 0.927 & 0.247 & 0.909 & 0.206 & 0.911 \\ \hline
  UNCCD-IOS & 0.571 & 0.925 & 0.833 & 0.873 & 0.000 & 0.926 & 1.000 & 0.807 & 0.100 & 0.810 & 0.258 & 0.838 & 0.300 & 0.913 & 0.848 & 0.903 & 0.989 & 0.827 & 0.288 & 0.832 \\ \hline
  LOF       & 0.000 & 0.985 & 0.667 & 0.985 & 0.778 & 0.618 & 1.000 & 0.793 & 0.033 & 0.938 & 0.161 & 0.919 & 0.600 & 0.930 & 0.370 & 0.985 & 0.409 & 0.958 & 0.031 & 0.973 \\ \hline
  DBSCAN    & 0.000 & 0.955 & 0.833 & 0.993 & 0.000 & 0.980 & 0.600 & 1.000 & 0.000 & 0.943 & 0.161 & 0.955 & 0.100 & 0.989 & 0.304 & 0.996 & 0.376 & 0.992 & 0.000 & 0.959 \\ \hline
  MST       & 0.429 & 0.866 & 0.500 & 0.718 & 0.778 & 0.662 & 0.700 & 0.756 & 0.367 & 0.695 & 0.774 & 0.437 & 0.600 & 0.782 & 0.652 & 0.553 & 0.892 & 0.668 & 0.553 & 0.672 \\ \hline
  ODIN      & 0.429 & 0.746 & 0.833 & 0.873 & 0.111 & 0.848 & 0.500 & 0.869 & 0.167 & 0.848 & 0.290 & 0.874 & 0.600 & 0.835 & 0.587 & 0.925 & 0.097 & 0.971 & 0.062 & 0.976 \\ \hline
  iForest   & 0.143 & 0.821 & 1.000 & 0.939 & 0.111 & 0.936 & 0.800 & 0.939 & 0.000 & 0.957 & 0.097 & 0.961 & 0.500 & 0.978 & 0.022 & 0.999 & 0.806 & 0.967 & 0.004 & 0.953 \\ \hline
\end{tabular}}
 \caption{The TPRs and TNRs of selected outlier detection methods on real-life datasets.}\label{Real_Data_Result_OS1}
\end{table}

\begin{table}[htb]
  \resizebox{\textwidth}{!}{\begin{tabular}{|c|c|c|c|c|c|c|c|c|c|c|c|c|c|c|c|c|c|c|c|c|}
  \hline
  \multirow{2}*{} & \multicolumn{2}{|c|}{hepatitis} & \multicolumn{2}{|c|}{lymph} & \multicolumn{2}{|c|}{glass} & \multicolumn{2}{|c|}{WBC}  & \multicolumn{2}{|c|}{vertebral} & \multicolumn{2}{|c|}{stamps} & \multicolumn{2}{|c|}{WDBC} & \multicolumn{2}{|c|}{vowels} & \multicolumn{2}{|c|}{thyroid} & \multicolumn{2}{|c|}{wilt} \\ \cline{2-21}
  & BA & $F_2$-score & BA & $F_2$-score & BA & $F_2$-score & BA & $F_2$-score & BA & $F_2$-score & BA & $F_2$-score & BA & $F_2$-score & BA & $F_2$-score & BA & $F_2$-score & BA & $F_2$-score \\ \hline
  U-MCCDs   & 0.583	& 0.263 & 0.600 & 0.222 & 0.681 & 0.257 & 0.506 & 0.168 & 0.555 & 0.335 & 0.511 & 0.072 & 0.607 & 0.170 & 0.664	& 0.196 & 0.530 & 0.117 & 0.696 & 0.336 \\ \hline
  SU-MCCDs  & 0.606 & 0.286 & 0.600 & 0.222 & 0.681 & 0.257 & 0.555 & 0.195 & 0.388 & 0.140 & 0.700 & 0.455 & 0.561 & 0.140 & 0.686	& 0.207 & 0.601 & 0.158 & 0.542 & 0.185 \\ \hline
  UN-MCCDs  & 0.686 & 0.446 & 0.572 & 0.189 & 0.493 & 0.116 & 0.487 & 0.159 & 0.474 & 0.036 & 0.648 & 0.381 & 0.591 & 0.146 & 0.771	& 0.263 & 0.550 & 0.126 & 0.519 & 0.117 \\ \hline
  SUN-MCCDs & 0.686 & 0.446 & 0.522 & 0.149 & 0.770 & 0.324 & 0.520 & 0.175 & 0.514 & 0.109 & 0.701 & 0.457 & 0.662 & 0.172 & 0.827	& 0.328 & 0.634 & 0.190 & 0.555 & 0.206 \\ \hline
  RKCCD-OOS & 0.433 & 0.000 & 0.613 & 0.227 & 0.503 & 0.122 & 0.525 & 0.135 & 0.519 & 0.139 & 0.561 & 0.230 & 0.753 & 0.302 & 0.613 & 0.221 & 0.582 & 0.161 & 0.509 & 0.092 \\ \hline
  RKCCD-IOS & 0.534 & 0.147 & 0.811 & 0.424 & 0.549 & 0.172 & 0.890 & 0.515 & 0.455 & 0.064 & 0.532 & 0.193 & 0.807 & 0.449 & 0.840 & 0.496 & 0.835 & 0.381 & 0.557 & 0.197 \\ \hline
  UNCCD-OOS & 0.576 & 0.256 & 0.765 & 0.347 & 0.525 & 0.137 & 0.548 & 0.156 & 0.502 & 0.104 & 0.540 & 0.181 & 0.787 & 0.380 & 0.670 & 0.310 & 0.578 & 0.160 & 0.559 & 0.178 \\ \hline
  UNCCD-IOS & 0.748 & 0.541 & 0.853 & 0.532 & 0.463 & 0.000 & 0.904 & 0.549 & 0.455 & 0.092 & 0.548 & 0.220 & 0.607 & 0.203 & 0.875 & 0.542 & 0.908 & 0.425 & 0.560 & 0.198 \\ \hline
  LOF       & 0.493 & 0.000 & 0.826 & 0.700 & 0.697 & 0.289 & 0.897 & 0.532 & 0.488 & 0.037 & 0.540 & 0.162 & 0.765 & 0.423 & 0.677	& 0.383 & 0.684 & 0.341 & 0.502 & 0.035 \\ \hline
  DBSCAN    & 0.478 & 0.000 & 0.913 & 0.833 & 0.490 & 0.000 & 0.800 & 0.652 & 0.471 & 0.000 & 0.557 & 0.178 & 0.697 & 0.435 & 0.650	& 0.343 & 0.684 & 0.401 & 0.673 & 0.381 \\ \hline
  MST       & 0.647 & 0.375 & 0.609 & 0.224 & 0.720 & 0.313 & 0.728 & 0.354 & 0.531 & 0.282 & 0.606 & 0.373 & 0.691 & 0.242 & 0.603	& 0.178 & 0.780 & 0.253 & 0.612 & 0.266 \\ \hline
  ODIN      & 0.587 & 0.313 & 0.853 & 0.532 & 0.480 & 0.074 & 0.684 & 0.342 & 0.507 & 0.159 & 0.582 & 0.262 & 0.717 & 0.286 & 0.756 & 0.427 & 0.534 & 0.093 & 0.519 & 0.069 \\ \hline
  iForest   & 0.482 & 0.122 & 0.965 & 0.750 & 0.524 & 0.100 & 0.869 & 0.656 & 0.479 & 0.000 & 0.529 & 0.108 & 0.739 & 0.472 & 0.510	& 0.027 & 0.887 & 0.664 & 0.479 & 0.004 \\ \hline
\end{tabular}}
 \caption{The BAs and $F_2$-scores of selected outlier detection methods on real-life datasets. In general, the two IOS not only outperform previous CCD-based methods but also surpass established methods.}\label{Real_Data_Result_OS2}
\end{table}

In the hepatitis dataset,
UNCCD-IOS delivers the highest $F_2$-score (0.541) among all the methods. 
The UN-MCCD and SUN-MCCD methods achieve the second-best performance with $F_2$-score equal to 0.446. 
In comparison, the MST method yields an $F_2$-score of 0.375. 
All the other methods deliver considerably lower TPRs, leading to worse performance.

For the Lymphography dataset, LOF, DBSCAN, and iForest achieve the best performance, 
with $F_2$-scores equal to 0.7, 0.833, and 0.750, respectively. 
The four CCD-based OSs deliver lower $F_2$-scores than these established methods due to substantially lower TPRs. 
Nevertheless, they exhibit solid improvement over the CCD-based methods. 
For example, the $F_2$-score of UNCCD-IOS is 0.532, substantially higher than that of the UN-MCCD method (0.189).

Consider the glass dataset, 
the SUN-MCCD method achieves the highest performance with an $F_2$-score of 0.324. 
LOF and the MST method deliver comparable $F_2$-scores of 0.289 and 0.313, respectively. 
The four OSs, along with DBSCAN, ODIN, and iForest, 
can only capture a small proportion of outliers, 
leading to poor performance. 
While the U-MCCD and SU-MCCD methods successfully identify all outliers, their TNRs are only 0.363.

The two ISOs get substantially higher $F_2$-scores (0.515 and 0.549) than the other CCD-based methods or OOSs under the WBC dataset, 
because both of them achieve a TPR of 1 while maintaining high TNRs. 
However, DBSCAN and iForest perform better, with $F_2$-scores being 0.652 and 0.656, respectively.
Other established methods perform much worse. 

The U-MCCD and MST methods obtain the highest $F_2$-score of 0.335 and 0.282 under the vertebral dataset. 
In contrast, most other methods can hardly identify any outliers, leading to much worse performance.

For the stamps dataset, 
the SU-MCCD, SUN-MCCD, and MST methods achieve the best $F_2$ Scores of 0.455, 0.457, and 0.373, respectively. 
The four OSs fail to deliver better performance due to low TPRs. 
Similarly, most other methods can barely distinguish the outliers from the regular points.

Under the WDBC dataset, 
RKCCD-IOS and UNCCD-OOS achieve $F_2$-scores of 0.449 and 0.380, respectively, 
exhibiting solid improvement over the previously proposed CCD-based methods. 
Meanwhile, they are comparable to other top-performing methods like LOF, DBSCAN, and iForest.

Under the vowels dataset, RKCCD-IOS and UNCCD-IOS achieve the best performance with $F_2$-scores of 0.496 and 0.542, 
outperforming the CCD-based methods by a large gap. 
ODIN and LOF provide second-tier performance with $F_2$-scores of 0.383 and 0.427, respectively. 
Except for the four, the performance of other methods is mediocre.

With the Thyroid dataset, 
iForest demonstrates superior performance with a $F_2$-score of 0.664, 
RKCCD-IOS, UNCCD-IOS, and DBSCAN achieve the next best results, 
obtaining $F_2$-scores of 0.381, 0.425, and 0.401, respectively. 
The other methods exhibit inferior performance, with $F_2$-scores around or below 0.3.

Lastly, for the Wilt dataset, 
DBSCAN and the U-MCCD method attain the highest $F_2$-scores of 0.381 and 0.336, respectively. 
All other methods show substantially weaker performance. 
The four OSs get TPRs below 0.4, 
leading to low $F_2$-scores, 
exhibiting limited improvement over the CCD-based methods.

In conclusion, while the two OOSs deliver comparable overall performance compared to the CCD-based methods, 
the two IOSs demonstrate significant improvement. 
They not only outperform previous CCD-based methods but also surpass established methods in most real-world datasets, 
making them highly competitive options for outlier detection. 

\section{Summary and Conclusion}

In this paper, 
we introduced two novel OSs based on CCDs: OOS and IOS. 
Both scores enhance the interpretability of outlier detection results. 
Both OSs employ graph-, density-, and distribution-based techniques, tailored to high-dimensional data with varying cluster shapes and intensities.
The newly proposed OSs can identify both global and local outliers and are invariant to data collinearity.
The IOS method, in particular, 
exhibits robustness to the masking problem, 
which is a challenge for many outlier detection techniques.  
Through extensive Monte Carlo simulations and real-life data examples, 
we demonstrated the efficacy of both OSs, 
showcasing their superior performance in comparison to existing CCD-based methods and other state-of-the-art approaches.

\section{Acknowledgements}
Most of the Monte Carlo simulations in this paper were completed in part with the computing resource provided by the Auburn University Easley Cluster. 
The authors are grateful to Art{\"u}r Manukyan for sharing the codes of KS-CCDs and RK-CCDs.

\section*{Declaration of generative AI and AI-assisted technologies in the writing process}
During the preparation of this work, the author used ChatGPT (OpenAI) to assist with drafting, rephrasing, and refining portions of the text for clarity and conciseness. After using this tool, the author thoroughly reviewed and edited the content to ensure accuracy, coherence, and alignment with the scholarly intent of the work, and takes full responsibility for the content of the publication.
 
\clearpage
\bibliographystyle{plain}
\bibliography{rf}

\end{document}